\documentclass[a4paper,fleqn]{cas-sc}

\usepackage[authoryear,longnamesfirst]{natbib}
\usepackage{amsmath}
\usepackage{amssymb}
\usepackage{booktabs} 
\usepackage{caption} 
\usepackage{subcaption} 
\usepackage{graphicx}
\usepackage{pgfplots}
\usepackage[all]{nowidow}
\usepackage[utf8]{inputenc}
\usepackage{tikz}
\usetikzlibrary{er,positioning,bayesnet}
\usepackage{multicol}
\usepackage{algpseudocode,algorithm,algorithmicx}
\usepackage{hyperref}
\usepackage[inline]{enumitem} 
\usepackage{subcaption}
\usepackage{multirow}
\usepackage{tabularx}
\usepackage[version=4]{mhchem}
\pgfplotsset{compat=1.18}
\def\tsc#1{\csdef{#1}{\textsc{\lowercase{#1}}\xspace}}
\tsc{WGM}
\tsc{QE}
\tsc{EP}
\tsc{PMS}
\tsc{BEC}
\tsc{DE}

\begin{document}
\let\WriteBookmarks\relax
\def\floatpagepagefraction{1}
\def\textpagefraction{.001}

\shorttitle{Enhancing Emergency Decision-making with Knowledge Graphs and Large Language Models}

\shortauthors{Minze Chen et~al.}

\title [mode = title]{Enhancing Emergency Decision-making with Knowledge Graphs and Large Language Models}                      
\tnotemark[1]

\tnotetext[1]{This work is supported in part by the National Key Research and Development Program of China under Grant No. 2022YFB3207704, National Natural Science Foundation of China under Grant No. 52302426, Youth Foundation of State Key Laboratory of Explosion Science and Technology under No. QNKT23-12.}


%
\author[1]{Minze Chen}[style=chinese,orcid=0000-0003-4925-1604]
\ead{chenmz22@bit.edu.cn}
\credit{Conceptualization of this study, Methodology, Resources, Software, Writing – original draft, Writing -
review \& editing, Data curation Validation, Visualization}
\affiliation[1]{organization={State key laboratory of Explosive Science and technology\& School of Mechachonical Engineering},
    addressline={Beijing Institute of Technology}, 
    postcode={100081}, 
    state={Beijing},
    country={China}}

\author[2]{Zhenxiang Tao}[style=chinese,orcid=0000-0001-5200-6234]
\affiliation[2]{organization={School of Emergency Management and Safety Engineering},
    addressline={China University of Mining and Technology-Beijing}, 
    postcode={100083}, 
    state={Beijing},
    country={China}}
\ead{ztao@cumtb.edu.cn}
\credit{Methodology, Resources, Software, Writing –review \& editing, Data curation Validation}

\author[3]{Weitong Tang}[%
   style=chinese,
   ] 
\credit{Conceptualization of this study, Methodology, Data curation}
\affiliation[3]{organization={Department of Engineering Physics},
    addressline={Tsinghua University}, 
    postcode={100084}, 
    state={Beijing},
    country={China}}

\author[4]{Tingxin Qin}[%
   style=chinese,
   ]
\ead{qintx@cnis.ac.cn}
\credit{Data curation, Resources, Funding acquisition}
\affiliation[4]{organization={China National Institute of Standardization}, 
    postcode={100088}, 
    state={Beijing},
    country={China}}

\author%
[3]
{Rui Yang}[%
   style=chinese,
   ]
\cormark[1]
\ead{ryang@tsinghua.edu.cn}
\credit{Conceptualization of this study, Methodology, Resources, Software, Writing –review \& editing, Visualization, Funding acquisition, Project administration, Supervision}

\author%
[1]
{Chunli Zhu}[%
   style=chinese, orcid=0000-0002-6109-9484
   ]
\cormark[1]
\ead{chunlizhu@bit.edu.cn}
\credit{Resources, Writing –review \& editing, Funding acquisition, Project administration, Supervision}

\cortext[cor1]{Corresponding author}


\begin{abstract}
Emergency management urgently requires comprehensive knowledge while having a high possibility to go beyond individuals' cognitive scope. Therefore, artificial intelligence(AI) supported decision-making under that circumstance is of vital importance. Recent emerging large language models (LLM) provide a new direction for enhancing targeted machine intelligence. However, the utilization of LLM directly would inevitably introduce unreliable output for its inherent issue of hallucination and poor reasoning skills. In this work, we develop a system called Enhancing Emergency decision-making with Knowledge Graph and LLM (E-KELL), which provides evidence-based decision-making in various emergency stages. The study constructs a structured emergency knowledge graph and guides LLMs to reason over it via a prompt chain. In real-world evaluations, E-KELL receives scores of 9.06, 9.09, 9.03, and 9.09 in comprehensibility, accuracy, conciseness, and instructiveness from a group of emergency commanders and firefighters, demonstrating a significant improvement across various situations compared to baseline models. This work introduces a novel approach to providing reliable emergency decision support.

\end{abstract}


\begin{highlights}
\item An accurate, reliable, and reasonable decision support system for emergencies, E-KELL.
\item Knowledge graphs constructed by structured Chinese emergency standards and regulations.
\item Automatically guiding large language models to reason over knowledge graphs.
\item The performance of E-KELL in various emergency conditions is well evaluated.
\end{highlights}


\maketitle

\section{Introduction}

Emergency Decision Support Systems (EDSS) are computer-based information systems that provide interactive support to help better decision-making under emergencies, such as floods, earthquakes, and pandemics (\cite{osatuyi2013temporal,levy2016advances,cremen2022developing,xu2021prediction,guler2020decision,deveci2022fuzzy}). Such emergencies can result in severe life and economic losses, with some disasters leading to cascading effects and additional technological hazards in a short period. By analyzing historical cases and emergency plans, EDSSs offer auto-generated recommendations, such as incident categorization, hazard evaluation, and disposal plan options (\cite{imran2020using, fertier2020new,imran2020using}). These systems mitigate the risk of decision misjudgment, especially in the case of rare or novel incidents. EDSSs also save valuable response time, crucial for rapidly worsening emergencies. However, current EDSSs typically offer information directly related to the incident, leaving emergency managers to interpret and make decisions themselves.

Large language models (LLM), such as ChatGPT, have gained widespread attention. They have the potential to play a significant role in the next generation of EDSS because of their advanced skills in understanding natural language, synthesizing information from diverse sources, and generating human-like responses. (\cite{jiang2022chatbot,RAY2023121,yang2021deep}). An LLM-driven EDSS application could understand users' needs and analyze them. Instead of obscure standards, users are provided direct guidance and engage in multi-turn conversations.

Despite the remarkable advancements achieved by LLMs, they share some limitations. Firstly, such models often lack a comprehensive understanding of the background knowledge associated with their generated text, which can lead to inconsistencies or inaccuracies in the output. As a result, LLMs sometimes generate false responses, such as incorrect facts, false character, or meaningless sentences, which is the so-called "hallucinations"(\cite{salvagno2023artificial}). Secondly, current LLMs' capacity for reasoning and establishing logical connections between distinct concepts is constrained, potentially resulting in disjointed responses (\cite{qi2023limitation}).

The limitations above become particularly critical in emergencies. Specifically, the hallucinations directly affect the effectiveness of decisions made. Meanwhile, EDSSs demand high interpretability and reference of every decision-making step, which helps decision-makers gather further information and avoid ethical and legal risks. Furthermore, the capacity to reason and establish logical connections between different concepts is crucial for identifying optimal courses of action. EDSS needs to provide decision support that considers both standard procedures and real-world conditions.

In addition, the most common basis for emergency response is regulations and standards published by government authorities. Among them, much information is stored in fragmented text forms, such as tabular and block diagrams. Current approaches for enhancing LLM by in-context learning often involve inputting text extracted from documents in PDF format. However, LLMs need help comprehending the intrinsic logical relationships between components in tables since they only receive fragmented text from these sources. This limitation highlights the need for enhancing the capabilities of LLMs in processing fragmented information from complex documents. Consequently, These drawbacks impede the extensive use of LLMs in emergency decision-making.

This paper presents the research and development behind a framework. Our work focuses on accurate and reliable decision support. The main contributions are as follows:

\begin{enumerate}
\itemsep=0pt
\item The paper develops a novel framework of EDSS based on KGs over an LLM application framework, which enables prompt feedback with interpretable guidance.
\item The study introduces an approach to structure Chinese emergency documents and guide LLM reason over KG by a prompt chain.
\item E-KELL is demonstrated to have superior decision support capabilities in various emergency conditions.
\end{enumerate}

In this paper, section 2 reviews the relevant literature on related works. Section 3 presents the methodology used in this research. Section 4 describes the implementation details of the proposed system, including system architecture, data preprocessing, and user interface design. Furthermore, in section 5, a case study is conducted to evaluate the system's effectiveness. Section 6 explores the significance of the research findings and potential outcomes. Section 7 concludes the paper and summarizes the key contributions of the research.

\section{Related work}

\subsection{Industry LLMs} 

Following the breakthrough progress of LLMs in the past year, numerous research teams and business companies have shifted their focus towards developing industry-specific LLMs, which can better adapt to specific tasks to meet the application demands across various industries (\cite{head2023large}). There are typically two methods to accomplish this task. The first is adjusting LLMs through \textbf{fine-tuning}, which involves retraining the pre-trained LLMs to adapt to the specific needs of a task or field (\cite{ding2023parameter}). Fine-tuning is a transfer learning method, adjusting the model parameters through relatively short training on a smaller dataset. The second method builds in-context-learning-based LLM application frameworks, which input both queries and related text to LLMs through a prompt manager (\cite{martino2023knowledge}).

The fine-tune-based industry LLMs exhibit rapid responses and superior generalization capabilities due to their enhanced adaptability to varying data distributions. Various studies successfully establish LLMs for specific industries through fine-tuning. \cite{liga2023fine} fine-tunes GPT-3 to easily classify legal and deontic rules on small amount of data. \cite{jiang2023health} introduce a method to tame LLM to be a doctor through leveraging real-world data from doctors in the supervised fine-tuned stage. \cite{singhal2023large} presents a benchmark combining existing medical question answering datasets with a new dataset of medical questions searched online and proposes a human evaluation framework. \cite{fan2023grammargpt} introduces an approach to explore LLMs' potential for native Chinese grammatical error correction. However, fine-tuning still fails to address the hallucinations effectively. 

Despite the rapid advancements in this field due to increased attention, Industry LLMs are not yet ready to replace domain experts, particularly in emergency response, healthcare, and law (\cite{thirunavukarasu2023large}). A key reason is the ethical and legal concerns that errors can provoke (\cite{mesko2023imperative}). Some researchers incorporate ethical considerations into the application framework (\cite{nussbaumer2023framework}). While such strategies can help mitigate potential risks, the fundamental solution lies in bolstering decision transparency and interpretability from a technical perspective. 

Compared to fine-tuning, prompt-based industry LLMs offer a better solution for enhancing the accuracy of responses. It can be easily understood as providing LLMs with tips containing the correct answers before they respond to queries. \cite{mansurova2023development} develops a system which allows LLMs to utilize blockchain domain data stored in vector databases, providing them with relevant information for their responses. \cite{huang2023dsqa} implements knowledge Q\&A in the military field by using template learning and template matching methods based on LLMs. \cite{yang2023counterfactual} introduces ChatGPT for data enhancement and combines it with label reset to construct counterfactual data in medical question and answering. \cite{jiang2023legal} introduces a simple prompting method to teach LLMs for legal judgment prediction.

In emergency management, \cite{goecks2023disasterresponsegpt} develops a prompt-based system with multi-round dialogue feedback. However, providing adequate context for these LLMs in practical application is a significant challenge. Indeed, a vast amount of knowledge in real life is unstructured, which makes it difficult for LLMs to comprehend and conduct further reasoning. To address that issue, some researchers consider unifying KGs into LLMs.

\subsection{Integration of KG and LLM}

LLMs often fall short of capturing factual knowledge for being black-box models. In contrast, KGs are structured knowledge models that explicitly store rich factual knowledge. Therefore, it is complementary to integrate LLMs and KGs and simultaneously leverage their advantages (\cite{10286238}). The current focus of existing research primarily lies in two aspects: applying LLMs to construct KGs (\cite{shi2023chatgraph,mehta2023enabling}) and analyzing KGs through LLMs (\cite{vassiliou2023isummary,choudhary2023complex}).

In the case of constructing KGs by LLMs, \cite{shi2023chatgraph} uses ChatGPT to transform knowledge into graphs for text classification. \cite{yao2023exploring} guides LLMs for triple classification and relation prediction in incomplete KGs. In disaster management, \cite{mehta2023enabling} constructs knowledge graphs and achieves knowledge discovery from natural disaster engineering datasets. These methods prove the superiority of KG construction and completion. 

As for the analytical aspect, knowledge graphs have proved to be a superior form for representing entities and their logical relationships, which are better for machines to understand. \cite{vassiliou2023isummary} develops a system to summarize the content of knowledge graphs.  \cite{choudhary2023complex} propose a novel approach integrating LLM for logical reasoning over KGs. They have achieved good results in various reasoning tasks but still need to complete automated processes. However, these knowledge analysis techniques for structured data are proven to have the potential to address the "hallucination" caused by LLM and enhance reasoning capability.

As we know, E-KELL stands as the first solution based on KG-enhanced LLM frameworks specially developed for emergencies. In contrast to the numerous industry-specific LLM fine-tuned models, E-KELL is a more suitable solution due to its heightened reliability. Furthermore, thanks to its structured knowledge repository and reasoning methods, E-KELL demonstrates an improved capacity to comprehend the emergency knowledge base and engage in logical reasoning.

\section{Methodology}

This section presents a framework for building an enhanced EDSS. Such an approach addresses providing reliable decision support in emergencies. In addition, this approach applies new technologies for text information mining and reasoning while ensuring interpretability as much as possible.

\begin{figure}
	\centering
		\includegraphics[width=13cm]{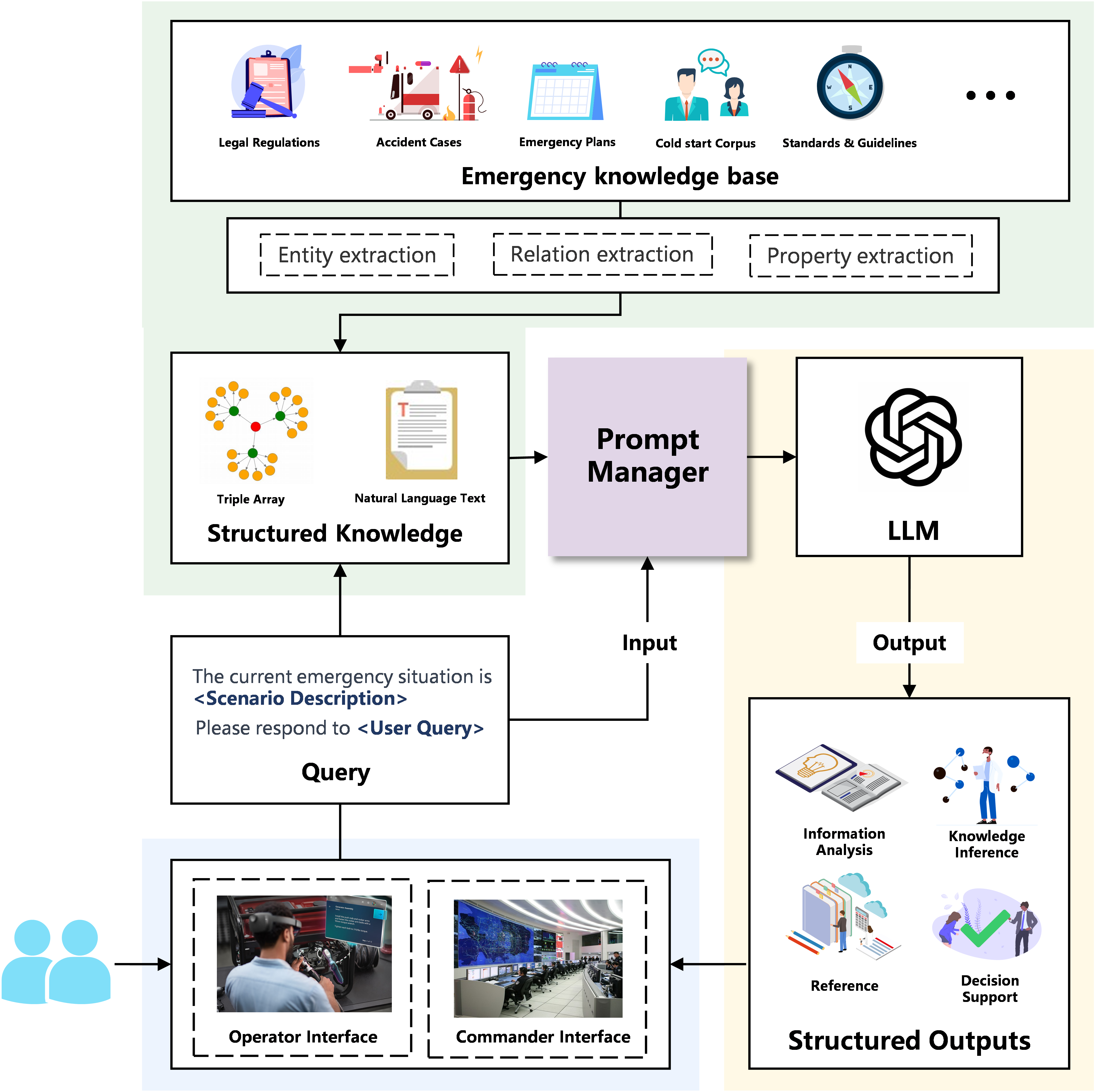}
	\caption{An overview of the proposed E-KELL framework}
	\label{FIG:1}
\end{figure}

Figure \ref{FIG:1} presents an overview of the proposed framework. It is elaborated and applied in the specific background of emergency decision support, which demands higher reliability and rapid responsiveness. However, the overall concept can be transferred and applied to other domains. In particular, it serves as a guide for developing an interactive knowledge base in any other area.

\subsection{Knowledge structuring} \label{sec:structure-learning}

The basis of emergency decision-making is relevant laws, accident cases, emergency plans, cold start language resources, standards, and guidelines. They have varying structures and styles since different government authorities and experts publish them. On one hand, they contain a significant amount of redundant information. On the other hand, transforming terms of documents into actionable guidelines requires high domain expertise. Extracting the logical meaning and rules from the text to construct a knowledge graph is beneficial for assisting machines in accurately understanding the main content of the documents.

Knowledge graph is a structured semantic knowledge repository. In a knowledge graph, knowledge is stored in the form of triples.

\begin{equation}
(h,r,t)\in{\mathcal{T}}
\end{equation}

where $\mathcal{T}$ is triple set. $\mathcal{E}$ and $\mathcal{R}$ represent entity set and relation set. $h\in{\mathcal{E}}$ is head entity. $t\in{\mathcal{E}}$ is tail entity.$r\in{\mathcal{R}}$ represents relation between head and entities.

In the graphs, nodes represent entities or entity attributes, and edges represent relations. Therefore, besides the typical triple as entity-relation-entity, entity-attribute-property value triples provide more information of entities. In terms of logical structure, a KG consists of a schema layer and a data layer. KG construction is either in a top-down or bottom-up manner. The former constructs the pattern layer first, which involves pre-defining the hierarchical relationships of entities, relations, and attributes. Then, specific data is populated into the schema layer. In this work, we construct the schema layer first because it is more suitable for domain-specific KGs for its predefined structure.

\begin{figure}
	\centering
		\includegraphics[width=14cm]  {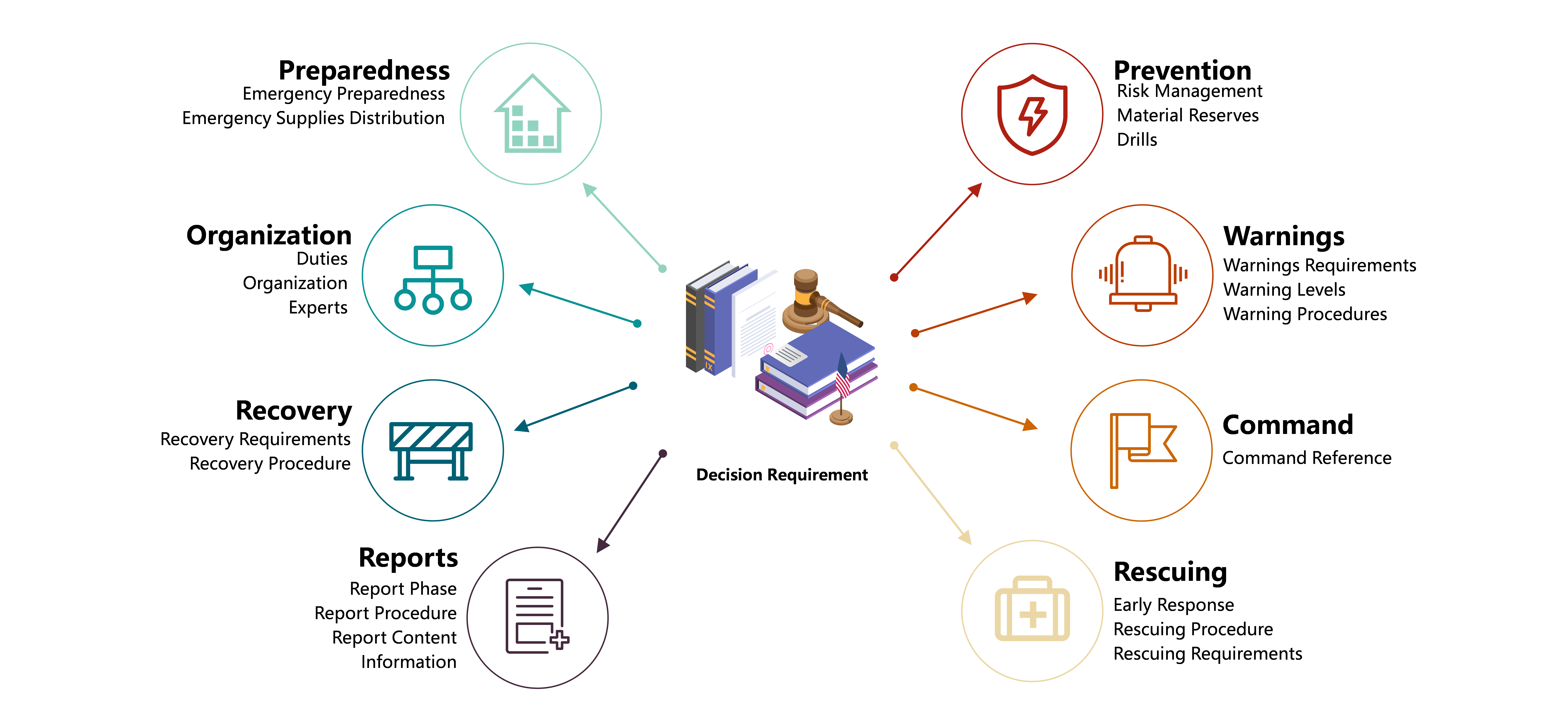}
	\caption{Decision demands classification and identification in emergencies}
	\label{FIG:2}
\end{figure}

Considering the decision response in emergencies, the decision demand is identified according to the common tasks and standard documents. It is categorized into eight primary classes and 22 subclasses, as shown in Figure \ref{FIG:2}.

The schema layer can be constructed based on these decision-making requirements. However, Converting documents into triples and adding them to the data layer is still a substantial amount of work. Current automatic methods are still immature as they need a comprehensive understanding of knowledge. As a result, A semi-automatic approach is employed to reduce the workload while striving to ensure the quality of knowledge graph construction as much as possible. First, we adopt an LLM-based approach to deconstruct triples from the text. We select authoritative emergency standards and policy documents as the data sources. These documents cover various emergencies and contain recommendations or requirements for emergency decision-making. We then extract important concepts and relations from these documents through natural language processing techniques to construct the schema of the KG. The schema includes defining entity types, relation types, and attributes.Based on the schema, we parse the selected documents and extract semantic triples for the KG. Each triple represents one piece of emergency knowledge in the subject-predicate-object format. The subjects and objects correspond to entity types in the schema, such as “emergency event,” “response action,” and “relief resource.” The predicates correspond to relation types, such as “leads to,” “requires,” “consumes.”. The knowledge graph represents the primary relationships within the textual documents. It serves as a knowledge repository, supporting emergency decision-making by providing references from multiple related standards. 

Besides the knowledge extraction, knowledge fusion and updates are primarily carried out manually. These two tasks require a precise understanding of entity meanings to merge them, which is difficult in emergencies, as similar situations may demand entirely different responses. Common automatic methods such as text clustering and embedding are not sufficiently accurate in our test. In addition, knowledge fusion over the fixed KG is also complete by experts manually. Subsequent updates mainly involve updating specific knowledge instead of the schema layer. For example, for revised standards, corresponding triples with content changes must be replaced promptly. Similarly, triples related to standards that have been abolished should be removed from the data layer.

\subsection{Logical reasoning} 

The constructed KG integrates heterogeneous knowledge from different standards and policy documents. It can reveal logical connections among various concepts and allows inference across documents. For example, a standard may specify that a "flood" event leads to a "power outage," and another document may recommend several "response actions" for a "power outage," such as "dispatching emergency repair crew." Through the KG, we can infer that these "response actions" may also apply to the "flood" event.

In addition to understanding the logical relationships within KGs, proactively logical reasoning on existing KGs is a challenging topic in artificial intelligence research. Current studies focus on creating diverse latent space geometric structures to transform logical relationships into geometric representations in space (\cite{le2023knowledge,pan2023logic}). The latest research proposes a promising strategy (\cite{choudhary2023complex}). It transforms complex logical structures into prompts, guiding LLMs to multi-step reason on the KG. However, the accurate expression of query logic in the research is provided beforehand, which is not practical in real-world applications. To address this issue, we introduce a module at the outset of Choudhary's framework, which automatically decomposes queries guided by the prompt. The module requires LLM to learn how to break down queries based on provided examples. We adopt the same logical expression as Choudhary, which encompasses four core first-order logical (FOL) operations: projection (p), intersection ($\wedge$), union ($\vee$), and negation ($\neg$) for querying the knowledge graph.

\begin{align}
&q_p\left[Q_p\right] \triangleq ? V_p:\left\{v_1, v_2, \ldots, v_k\right\} \subseteq E \exists a_1&\\
&q_{\wedge}\left[Q_{\wedge}\right] \triangleq ? V_{\wedge}:\left\{v_1, v_2, \ldots, v_k\right\} \subseteq E \exists a_1 \wedge a_2 \wedge \ldots \wedge a_i&\\
&q_{\vee}\left[Q_{\vee}\right] \triangleq ? V_{\vee}:\left\{v_1, v_2, \ldots, v_k\right\} \subseteq E \exists a_1 \vee a_2 \vee \ldots \vee a_i&\\
&q_{\neg}\left[Q_{\neg}\right] \triangleq ? V_{\neg}:\left\{v_1, v_2, \ldots, v_k\right\} \subseteq E \exists \neg a_1&
\end{align}

Where $q_p$, $q_{\wedge}$, $q_{\vee}$, and $q_{\neg}$ represent projection, intersection, union, and negation queries. $Q$ is components in queries which can be represented by $\left(e_1, r_1\right)$ or $\left\{\left(e_1, r_1\right),\left(e_2, r_2\right), \ldots,\left(e_i, r_i\right)\right\}$. $V$ is the corresponding result. $E$ is the entity set while $a_i$ is a Boolean indicator, which will be 1 if $e_i$ is connected to $v_i$ by relation $r_i$, 0 otherwise (\cite{choudhary2023complex}).

The method above can transform queries into corresponding logical expressions. Subsequently, based on these expressions, predefined prompt templates are invoked to guide the LLM in a chain. Through testing, it is found that, compared to the approach in Choudhary's study, where sub-prompts of complex logic are encapsulated into a single paragraph provided to the LLM, a more effective approach is to input the sub-prompts step by step into the LLM for chain guidance. This approach also aligns with the consensus prompt engineering principle: Specify the steps to complete a task.

Additionally, to achieve effective query results, LLM does not require access to the entire KG. It only needs to access the most relevant portions of the KG that may contain answer-related information. It can be accomplished by relaxing the similarity requirements in retrieval or by neighborhood expansion based on retrieval results. These segments of the knowledge graph serve as the foundation for LLM's reasoning. The $k$-level neighborhood expansion is defined as:

\begin{align}
\mathcal{N}_1\left(q_\tau\left[Q_\tau\right]\right) & =\left\{(h, r, t):\left(h \in E_\tau^1\right),\left(r \in R_\tau^1\right),\left(t \in E_\tau^1\right)\right\} \\
E_\tau^k & =\left\{h, t:(h, r, t) \in \mathcal{N}_{k-1}\left(q_\tau\left[Q_\tau\right]\right)\right\}, \\
\quad R_\tau^k & =\left\{r:(h, r, t) \in \mathcal{N}_{k-1}\left(q_\tau\left[Q_\tau\right]\right)\right\} \\
\mathcal{N}_k\left(q_\tau\left[Q_\tau\right]\right) & =\left\{(h, r, t):\left(h \in E_\tau^k\right),\left(r \in R_\tau^k\right),\left(t \in E_\tau^k\right)\right\}
\end{align}
where $\tau$ is the query type. $E_\tau^1$ and $R_\tau^1$ are the set of entities and relations in query $Q_\tau$.

\begin{figure}
	\centering
		\includegraphics[width=14cm]  {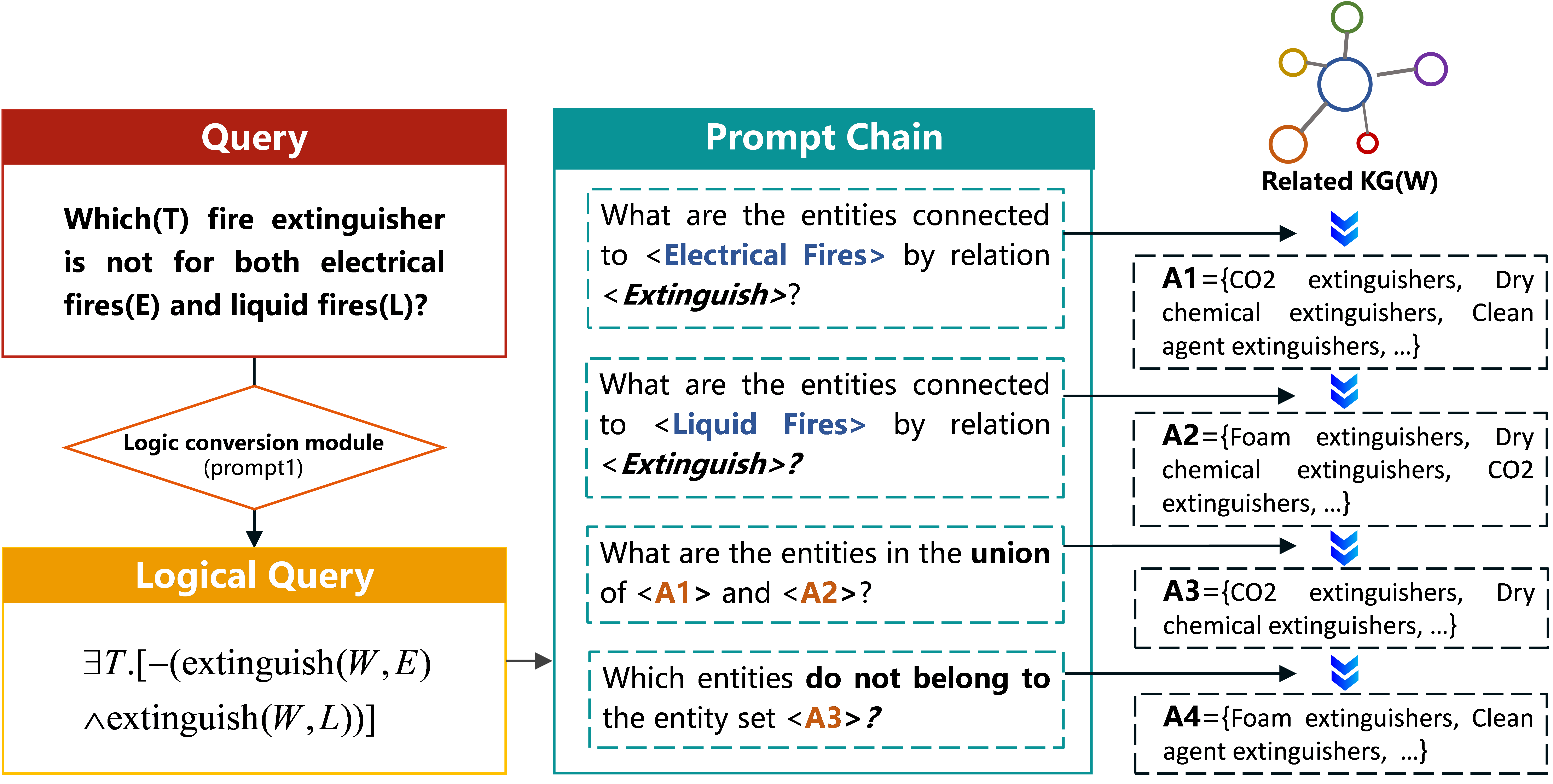}
	\caption{Example of Transforming a Query into a Prompt Chain and Guiding LLM Reasoning over the related KG}
	\label{FIG:3}
\end{figure}

Figure \ref{FIG:3} illustrates an example query and its automated process behind. The query is input into a prompt template. A LLM is tasked with transforming this example into a logical query, a pre-determined expression. Subsequently, in the following automatic prompt, we ask the LLM to convert the logical query into a chain of prompts. This chain encapsulates the stepwise logical information of the query. The relevant KG and the components within the chain of prompts are sequentially input into the LLM. Accordingly, the LLM navigates through the KG stepwise, based on the prompts, until it locates the set of triples required to generate the responses.

\section{System implementation}

\begin{figure}
	\centering
		\includegraphics[width=14cm]  {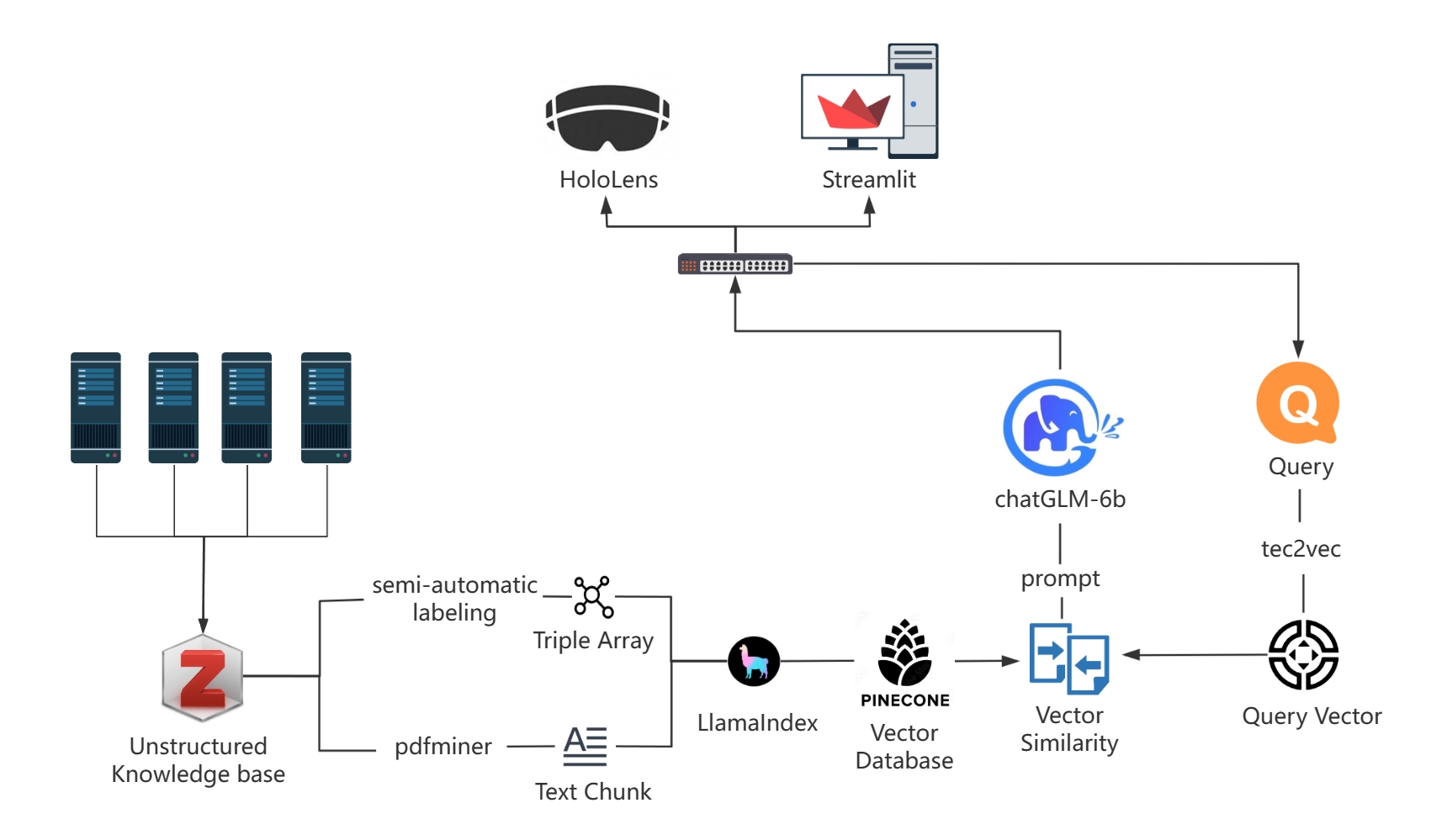}
	\caption{Softwares and packages used to implement the prototype}
	\label{FIG:4}
\end{figure}

This section discusses the implementation of the project prototype. The system prototype focuses on providing a practical response to hazardous chemical leakage. The softwares and packages used in the proposed system are illustrated in Figure \ref{FIG:4}. After gathering and structuring relevant information on hazardous chemical leakage, the processed results are stored in a vector database. The query module converts user queries into vectors and performs similarity matching with the structured knowledge in the vector database to extract the most relevant knowledge segments. Afterward, LLM is guide to reason within the provided KG segments by prompts. The relevant knowledge and customized prompts are organized in a prompt template pre-provided. Additionally, we develop two sets of user interfaces, one for decision centers and another for frontline operators, with the latter utilizing a hands-free interactive interface based on mixed reality (MR).

It is worth emphasizing that the documents essential for emergency decision support are notably for specific regions. The response-generate procedures must align with local laws and regulations, and the intended users are native speakers of the local language. Consequently, all pertinent standards and user interfaces compiled in this study are in Simplified Chinese. Considering the accessibility, we have translated the actual documents, interfaces, and prompts from Chinese into English, with no impact on the system's development or practical application.

This study focuses on the typical emergency scenario involving hazardous chemical leaks and aims to compile pertinent data for decision-making support. We source relevant materials from official Chinese government websites, legal databases, and similar sources. Documents must meet the criteria of being timely, relevant, and representative to be considered. A third-party OCR tool is employed for text mining, enabling text conversion into a computer-readable format suitable for bulk processing. The final compilation includes two legal regulations and ten standard specifications, along with other highly relevant documents, such as quick reference materials for hazardous chemicals. Using a structured framework encompassing entities, relationships, and attributes, we dissect the selected documents and extracted semantic triple arrays to populate the knowledge graph. Each triple array strictly follows the Subject-Predicate-Object format, representing essential emergency knowledge. The subject and object correspond to entity types within the structure, such as "emergency event," "response action," and "rescue resources." In contrast, the predicate corresponds to relationship types like "causes," "requires," and "consumes." These triples comprehensively capture the fundamental relationships present in the text. We manually disassemble a subset of content as examples, guiding the LLM to construct triples from the remaining extensive text by the prompt, as shown in Appendix A. The extensive triples obtained undergo manual fine-tuning to create the final KG. 

The method is employed exclusively for constructing a coarse KG. A significant reason is the need to comprehensively consider multiple standards and regulations when providing decision support for emergency events. It necessitates the LLM's capacity to retain an extensive volume of textual information. Given the limitations of existing LLMs, a group of volunteers with expertise in hazardous chemical leakage is engaged to revise the preliminary graph generated by the model. These volunteers are supposed to filter out redundant information and introduce cross-text triples. Consequently, a KG comprising 2264 triples is ultimately constructed. 

It is necessary to convert the emergency KG into vectors to retrieve the knowledge base quickly. In this study, our system employs the Llama index's vectorization module to embed the existing KG. Llama index is a flexible data framework that connects custom data sources to LLMs, providing a convenient interface for vector databases, which aims to feedback relevant KG sections for LLM to reason.

When users input descriptions of emergencies and queries into the system, they are represented as a vector matrix through text2vec. After similarity comparisons with the vector database, the system returns the most relevant portions of the KG along with their corresponding original texts from standards and regulations through reasoning. The queries and related knowledge are processed as predefined prompts in the prompt manager and sent to the LLM. The prompt templates used are provided in Appendix A. It is essential to enable LLM to reason and analyze based on the provided knowledge rather than its database, which effectively reduces the possibility of the LLM generating hallucinations.

Considering the data security of emergency knowledge and common interactive language, the system presented in this paper utilizes ChatGLM-6B, an open bilingual LLM based on the General Language Model (GLM) framework. We deployed the system locally on an NVIDIA A100 graphics card.

\begin{figure}
	\centering
		\includegraphics[width=12cm]  {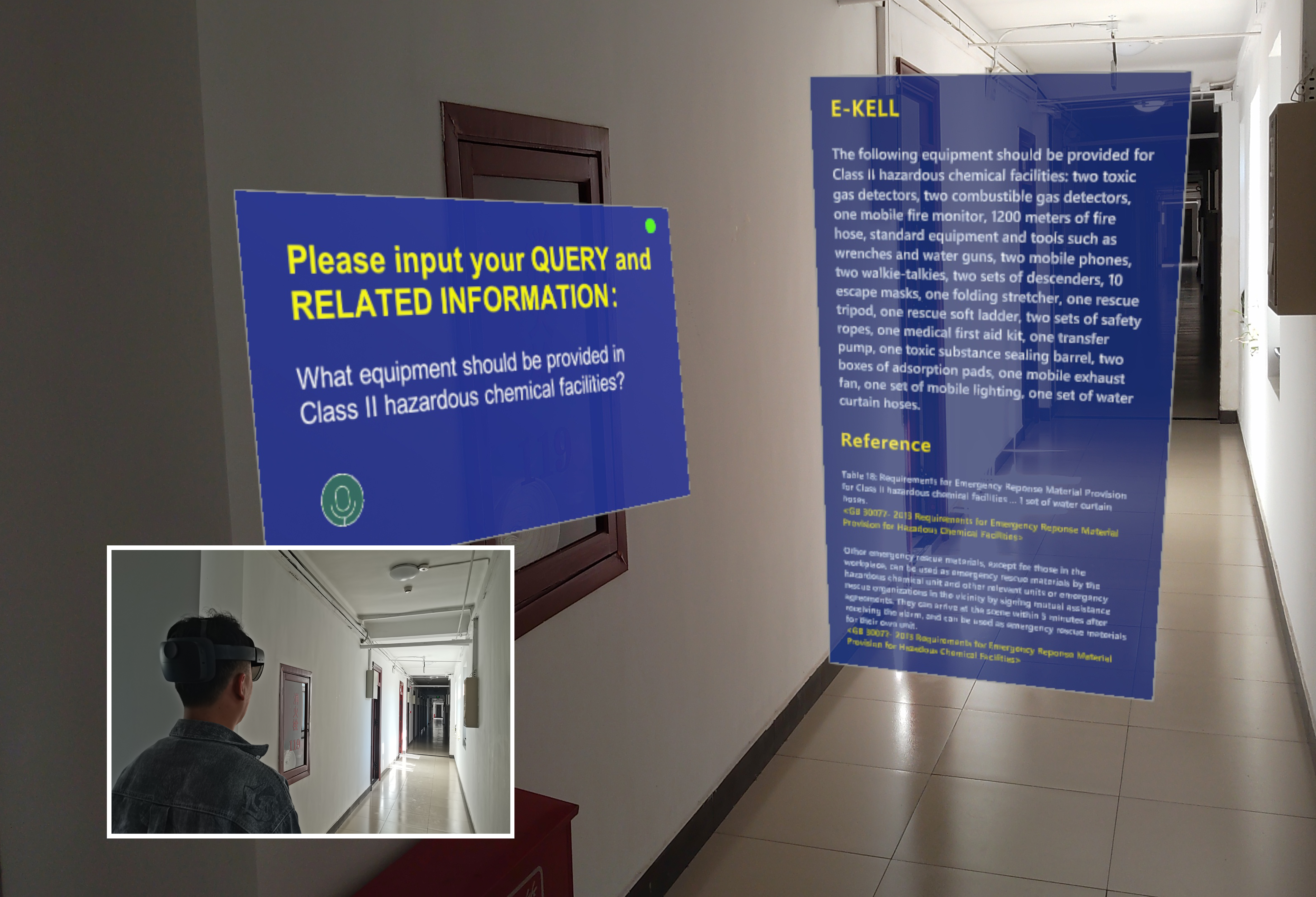}
	\caption{ (Main) A screenshot of the MR interface for the frontline operator. (Inset) A user sees the interface through the headset and engages in hands-free interaction}
	\label{FIG:5}
\end{figure}

Additionally, We design two user interfaces for the system, targeting commands and frontline operators. The proposed interfaces are available on both PC and MR platforms. The PC interface is explicitly designed for commands, which are responsible for making strategic decisions during emergencies. It provides an easy-to-navigate platform based on Streamlit. The system promptly displays relevant replies, along with the corresponding reference standards and regulatory items that form the basis of these replies. It aids commands in making well-informed decisions based on established guidelines. The MR interface is developed for frontline operators who require hands-free interaction during emergencies. As shown in Figure \ref{FIG:5}, the MR interface incorporates a voice-command feature, allowing users to ask questions and get answers without manual input. The interfaces can be moved to any spatial location to avoid obstructing the operator's work. It empowers them to concentrate on their duties while accessing vital information from the E-KELL. 

\section{Case study}

\subsection{Use case}

In this study, we selected ten queries, each representing decision points in emergency management of hazardous chemical leakage incidents, which emphasize methods rather than conclusions. They respectively represent decisions at different stages of emergency management for hazardous chemical leakage, including Material Reserves, On-site Command, Hazard Source Identification, Accident Investigation, Fire Extinguishment, Personal Protective Equipment, Rescuing, Isolation, Hazard Source Management, and Report Compilation. These queries were input into chatGLM-6b, GPT-3.5, and our developed system, E-KELL. The selection of these two baselines is motivated by chatGLM-6b being the foundational model for our system and GPT-3.5 being one of the most widely utilized LLMs in current applications. All prompts and corresponding responses are translated into English and presented in Appendix B.

Taking Q1 from Appendix B as an example, a typical query related to Material Reserves necessitates as precise responses as possible. By comparing the three responses, it becomes evident that incorporating a local knowledge base into LLMs can provide explicit guidance. The direct response from chatGLM-6b contains inconsistent information with the standard requirement. Compared to ChatGLM-6b, GPT-3.5's response is more detailed but still contains a large amount of redundant information. It increases the difficulty for users to obtain critical information while concise, direct, and quantifiable support is crucial for decision-making, especially in emergencies.

\begin{figure}
	\centering
		\includegraphics[width=14cm]  {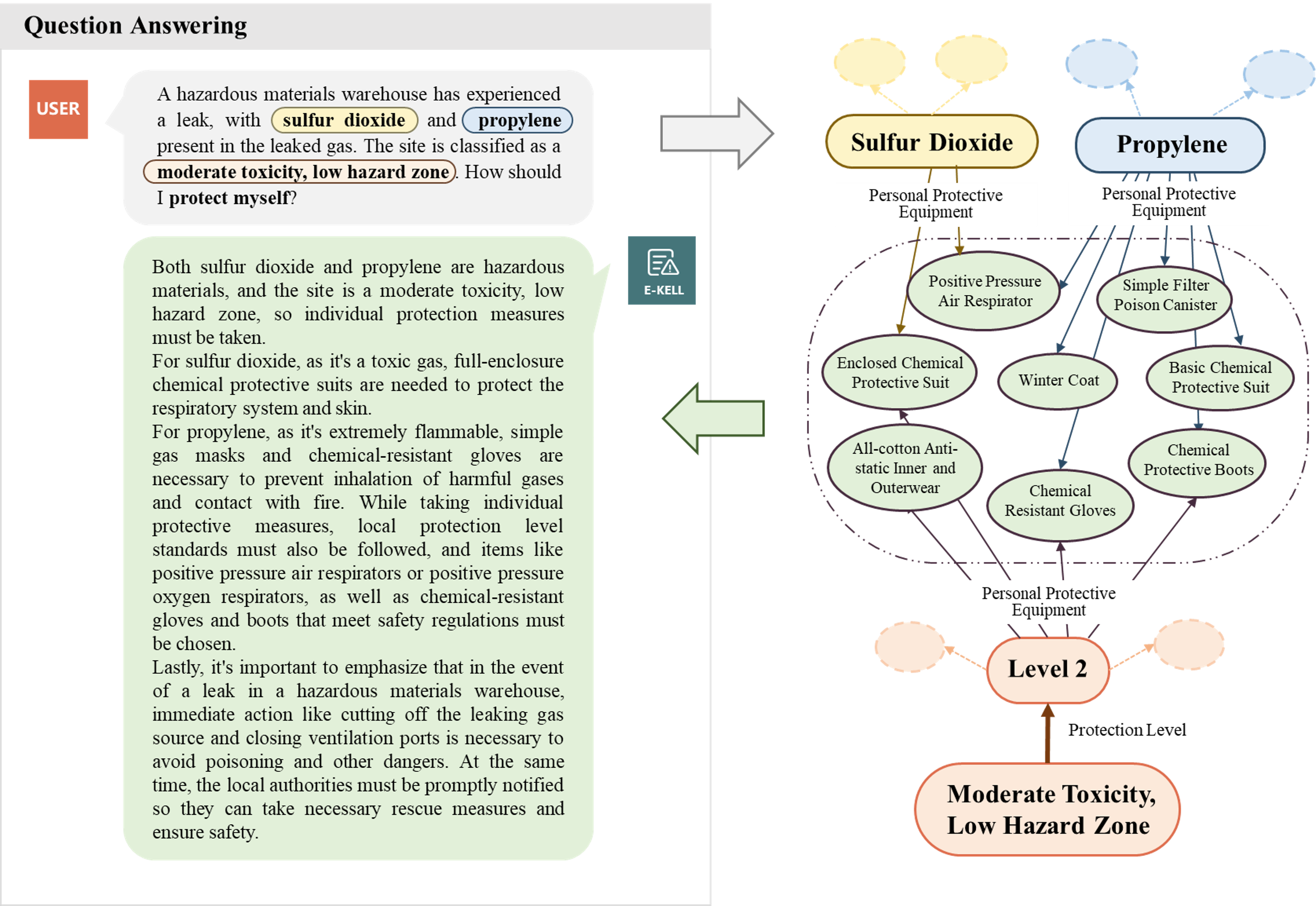}
	\caption{A typical query about Personal Protective Equipment, the response, and the underlying KG indexing}
	\label{FIG:6}
\end{figure}

Another example is illustrated in Figure \ref{FIG:6}. It represents a typical decision requirement about Personal Protective Equipment, which holds value for frontline operators.
The personal protective equipment required in hazardous material leaks requires considering established regulations and on-site conditions.

Figure \ref{FIG:6} also demonstrates the reasoning process in the LLM within the KG in E-KELL. The query necessitates individual protective requirements for situations involving "Sulfur dioxide," "Acetylene," and "Moderate toxicity, low hazard zone." Consequently, the system indexes the intersection of these factors. Through domain expansion retrieval, it extends from the "Moderate toxicity, low hazard zone" to "Level 2" through the relation "Protection level." Subsequently, it retrieves the protection standards required for that protection level, compiling an operational guide containing all necessary information.

In contrast, results generated directly by chatGLM-6b and GPT-3.5 exhibit low operability. They need help comprehending the specific emergency situations required and their precise interrelations.

\subsection{Evaluation}
To assess the system's performance, we evaluate the objective attributes of responses to ten selected questions across three models as shown in Table \ref{tbl1}. "Grammatically correct" denotes the absence of Chinese writing or grammar mistakes. "Factually correct" signifies the absence of evident factual inaccuracies. "In compliance with standards/regulations" means no content deviates from established standards, regulations, or guidelines which ensures that decision-makers would not violate existing documents when following these responses.

\begin{table}[width=.9\linewidth,cols=4,pos=h]
\caption{Objective Attribute Scores of Use Cases}\label{tbl1}
\begin{tabular*}{\tblwidth}{@{} LCCC@{} }
\toprule
 & E-KELL & ChatGLM-6b & GPT-3.5 \\
\midrule
Grammatically correct     & 10/10     & 10/10 & 10/10 \\
Factually correct & 10/10              & 8/10          & 9/10 \\
In compliance with standards/regulations    & 10/10              & 4/10          & 7/10 \\
\bottomrule
\end{tabular*}
\end{table}

The results indicate that E-KELL furnishes decision support in the provided examples, devoid of grammatical errors or Factual errors. Additionally, it does not violate any standards, regulations, or guidelines since it generates responses based on the information within the knowledge base rather than independently. Conversely, the other two models exhibit some factual errors and instances of non-compliance with existing standards and regulations. It is principally due to the "hallucination," as they both generate factual errors when delineating the chemical formula. They each represent the chemical formula of propylene as \ce{C3H8O} and \ce{C3H4}, whereas the correct formula is \ce{C3H6}. In another example (Q8), isolating an area with a hazardous chemical leak requires assessing the wind direction before isolation. If the response generated by chatGLM-6b is followed, the isolation would violate standards and result in severe consequences.

However, An outstanding EDSS, in addition to meeting the basic requirements mentioned above, also has additional subjective attributes, which can not quantify. More specifically, superfluous or redundant information can impair efficiency, a factor of paramount importance in emergencies. Consequently, it becomes imperative to gather information that is not only concise and accurate but also highly instructive. Furthermore, emergency decision support places a strong emphasis on a subjective, user-centric approach to system engineering. Hence, we recruit an expert team comprising 14 frontline firefighters and five emergency commanders, totaling 19 volunteers. From the total of 30 responses to the ten queries above, we randomly select responses and ask volunteers to score them on a scale of 1-10 regarding comprehensibility, accuracy, conciseness, and instructional value. The experts are kept from the source of each model's responses. The average scores for each model across all queries are outlined in the Table \ref{tbl2}.

\begin{table}[width=.9\linewidth,cols=5,pos=h]
\caption{Expert Scores for Use Case Responses}\label{tbl2}
\begin{tabular*}{\tblwidth}{@{} LCCCC@{} }
\toprule
 & Comprehensibility & Accuracy & Conciseness   & Instructiveness\\
\midrule
E-KELL     & \textbf{9.06}     & \textbf{9.09} & \textbf{9.03} & \textbf{9.06} \\
ChatGLM-6b & 8.57              & 8.54          & 8.80          & 8.69 \\
GPT-3.5    & 8.78              & 8.82          & 8.62          & 8.84 \\
\bottomrule
\end{tabular*}
\end{table}

The mean scores across four metrics for E-KELL are not just superior to its baseline model, chatGLM-6b, but also surpass the commonly used GPT-3.5. These rating outcomes corroborate the superiority of E-KELL. Compared to ChatGLM-6b, GPT-3.5 excels in comprehensibility, accuracy, and instructiveness. For example, in Q5, the response generated by GPT-3.5 mentions that water should not be used to extinguish a fire caused by propylene because propylene can react with water, leading to combustion and exacerbating the fire, which is emphasized in hazardous materials handling standards. The response from chatGLM-6b does not mention this point. However, GPT-3.5 exhibits a drawback regarding excessive redundancy, leading to poor conciseness. This shortcoming may hinder emergency managers from promptly assimilating critical information and making informed decisions.

\section{Discussion}

Emergency decision support demands high response speed, intelligence, and reliability. Therefore, LLMs are practical tools for the next generation of EDSSs. However, the lack of "hallucination" in LLMs poses a critical constraint within emergency management. In this study, incorporating the KG technology into the framework of LLM applications is anticipated to enhance the reliability and effectiveness of the response. It will expedite the deployment of LLMs in the emergency industry. In this section, we will discuss the issues and prospects of E-KELL application.

The quality of prompt-based LLM application responses depends on the prompts. Therefore, it is necessary to adjust prompt templates and continuously optimize them based on real-world situations during application. In addition, it is found in our test that many commonly used methods for automatically labeling KGs need to yield satisfactory results for non-structured emergency knowledge. Consequently, constructing a structured knowledge base still heavily relies on manual labeling, resulting in high costs, resources, and a limitation on logical reasoning. In the Evaluation section, it is revealed that existing LLMs generate responses with common-sense errors in fields such as chemistry. Hence it is necessary to enhance the system with a substantial amount of common sense to reduce the reliance of LLMs on unreliable databases. In addition to accessible knowledge, a determinant for emergency decision support is real-time data, such as the quantity of deployable resources and certain parameters from on-site sensors. This necessitates providing E-KELL with real-time data interfaces.

Requesting LLMs to confirm knowledge and self-check with prompts is highly beneficial in mitigating "hallucination " and enhancing reliability, as demonstrated in the study (\cite{peng2023check}. This approach is well-founded, as in real-world workflows, repetitive verification is a common method for ensuring the accuracy of results. Some LLM frameworks' prompt templates also include confirming knowledge. Therefore, continuous self-check within the information flow is essential when employing E-KELL. An outlook is to explore the application of AI agent technology to enhance the E-KELL system. AI agents enable the utilization of various role-based AI agents to accomplish complex tasks. Local KG, web search engines, and other tools are provided to these agents. This new application framework holds promise for further improving the reliability and decision-making capabilities of E-KELL.

Moreover, it is crucial to enable E-KELL to comprehend multimodal information. For instance, images provide a direct means of acquiring real-time emergency scene information. On a more superficial note, integrating an image understanding module at the system's input end to convert images into natural language descriptions fed to LLM is a feasible approach. From an output perspective, instead of decision support text, a more practical approach involves utilizing mixed reality to visualize decision guidance through holographic image animations in real time. It is simple enough, as LLM can output the structured outputs required. These outputs can serve as data inputs to the mixed reality front end. This approach can empower decision-makers by offering an intuitive and interactive feedback mechanism.

\section{Conclusion}

In this paper, we present E-KELL, a framework and developed system for enhancing emergency decision support using KGs and LLMs. Our approach involves structuring and storing available emergency-related text in a standardized knowledge base. User query is inputted and serves as the basis for reasoning within the knowledge graph. The system can assist emergency managers in decision support within standards and regulations. Through our case study and evaluation, we demonstrate the significant superiority of E-KELL in various emergency scenarios. Additionally, we discuss some issues and outlooks of application. 

The main limitation of the study is the limited data source, which significantly affects the reasoning quality of E-KELL. Additionally, our prompt design needs further testing. Besides the continuous system optimization, our future work involves the development of an AI agent collaboration framework. We plan to consult real-world emergency decision-making organizations, building an automated workflow including data retrieval, data review, and comprehensive problem-solving.

\appendix
\section{Prompt Templates}

The prompt templates for queries with are provided in Table \ref{tbl3}, respectively.

\begin{center}
\begin{table}[width=.9\linewidth,cols=2,pos=h]
\caption{Full Prompt Templates of Queries}\label{tbl3}
  \begin{tabular}{p{0.15\linewidth} p{0.7\linewidth}}
    \toprule
    \makebox[0.15\textwidth][l]{Task} & \makebox[0.7\textwidth][l]{Template for Full Prompts}\\
    \midrule
    KG Construction                         & You are a networked intelligence helping a human track knowledge triples, about all relevant people, things, concepts, etc. and integrating them with your knowledge stored within your weights, as well as that stored in a knowledge graph. Extract all of the knowledge triples from the text. A knowledge triple is a clause that contains a subject, a predicate, and an object. The subject is the entity being described, the predicate is the property of the subject that is being described, and the object is the value of the property. thinking in English, but please keep the original language. 
    
    EXAMPLE

    INPUT: Ammonia is an irritating gas that can form explosive mixtures with air; inhalation can cause toxic pulmonary edema. It can cause eye, skin, and respiratory tract burns; if the gas leak cannot be shut off, the flame at the leak site should not be extinguished; when handling liquid ammonia, wear cold-resistant clothing.

    OUTPUT: (Ammonia, is, irritating gas) <|> (Ammonia, form, explosive mixtures with air) <|> (Ammonia, cause, toxic pulmonary edema) <|> (Ammonia, cause, eye burns) <|> (Ammonia, cause, skin burns) <|> (Ammonia, cause, respiratory tract burns) <|> (Handling liquid ammonia, require ,wearing cold-resistant clothing) 

    END OF EXAMPLE

    INPUT: "\{text\}"

    OUTPUT:""       \\
    \midrule
    Decision Support     &    There are the following triple arrays representing relationships between entities:

    "\{context\_str\}"

    Based on the above information, answer the following questions: "\{query\_str\}"
    
    Please refrain from using knowledge beyond the above information.    \\
    \midrule
    Logical Reasoning (Projection)  & What are the entities connected to "\{entity\}" by relation "\{relation\}"   \\
    \midrule
    Logical Reasoning (Intersection)  & What are the entities in the intersection of "\{entities1\}" and "\{entities2\}"   \\
    \midrule
    Logical Reasoning (Union)  & What are the entities in the union of "\{entities1\}" and "\{entities2\}"  \\
    \midrule
    Logical Reasoning (Negation)  & Which entities do not belong to the entity set "\{entities\}" \\
    \bottomrule
  \end{tabular}
  \end{table}
\end{center}

\section{Prompts and responses for evaluation}

\begin{enumerate}
\itemsep=0pt
\item \textbf{Q1: (Material Reserves)}What supporting equipment should be provided in Class II hazardous chemical facilities?
    \begin{itemize} 
    \item \textbf{E-KELL:} The following equipment should be provided for Class II hazardous chemical units: 
    
    Two toxic gas detectors;

    Two combustible gas detectors;

    One mobile fire monitor; 

    1200 meters of fire hose;

    Standard equipment and tools such as wrenches and water guns;

    Two mobile phones;

    Two walkie-talkies;

    Two sets of descenders;

    Ten escape masks;

    One folding stretcher;

    One rescue tripod;

    One rescue soft ladder;

    Two sets of safety ropes;

    One medical first aid kit;

    One transfer pump;

    One toxic substance sealing barrel;

    Two boxes of adsorption pads;

    One mobile exhaust fan;

    One set of mobile lighting;

    One set of water curtain hoses.
    
    \item \textbf{ChatGLM-6b:} Class II hazardous chemical units should be equipped with the following supporting equipment: a safety monitoring system to monitor the use and safety of chemicals and promptly detect and warn of safety risks. Leak detection instruments to detect chemical leaks and take timely measures to prevent harm to the environment and personnel. Firefighting equipment to respond to emergencies such as chemical fires, providing timely and effective rescue services. Laboratory equipment for conducting chemical research and testing to ensure accurate and reliable experimental data. Safety training equipment for providing safety training and education to chemical users, enhancing their safety awareness and operational capabilities.
    
    \item \textbf{GPT-3.5:} Class II hazardous chemicals have dangerous properties such as combustibility, toxicity, corrosiveness, and irritability. Their storage, handling, and use require special safety measures and supporting equipment. Depending on the specific circumstances, units handling Class II hazardous chemicals may need different supporting equipment to ensure safe production and environmental protection. Here are some supporting equipment that might be needed:
    
    Safety storage equipment: Includes fire-resistant, explosion-proof, and corrosion-resistant storage devices, such as chemical storage tanks, tank trucks, storage cabinets, etc.
    
    Protective equipment: Provides necessary personal protective equipment for workers, such as protective glasses, respiratory protectors, protective clothing, gloves, etc., to reduce the risk of exposure to hazardous chemicals.
    
    Leakage handling equipment: Includes leak capture equipment, emission processing equipment, etc., used to swiftly handle leaks when a chemical leakage accident occurs to prevent diffusion and environmental pollution.
    
    Ventilation equipment: Provides indoor ventilation systems to ensure indoor air circulation, prevent the accumulation of hazardous gases, and maintain indoor air quality.
    
    Firefighting equipment: Equips automatic fire alarm systems, fire extinguishers, fire hydrants, and other firefighting equipment, as well as emergency evacuation channels and firefighting materials, to ensure that measures can be taken quickly in emergencies such as fires.
    
    Monitoring instruments: Equips monitoring instruments to monitor parameters such as the concentration of harmful gases, temperature, and humidity in indoor air, ensuring a safe working environment.
    
    Emergency rescue equipment: Equips first aid kits, eye washers, emergency beds, etc., and trains relevant personnel for emergency rescue.
    
    Workbenches and equipment: For specific chemical operations, special workbenches, equipment, and tools may be needed to ensure the safety of operators and the accuracy of operations.
    
    Electrical equipment: Includes explosion-proof electrical equipment to ensure the electrical system does not cause fires or explosions.
    
    Training equipment: Equips training materials drill equipment, and provides safety operation training and emergency drills for employees.
    
    It should be noted that the specific equipment requirements for different types of units handling Class II hazardous chemicals may vary. The specific equipment should be determined based on local laws and safety regulations, as well as the nature, storage, and usage of the chemicals. Units should consider safety, environmental protection, and emergency response capabilities when configuring supporting equipment.

    \end{itemize}
\item \textbf{Q2: (On-site Command)}In the event of hazardous chemical leaks, what procedures should emergency responders follow for protection and entering the scene for rescue?
    \begin{itemize} 
    \item \textbf{E-KELL:} In the event of a hazardous chemical leakage accident, emergency responders should follow the below-mentioned process for protection and rescue:

Protective Measures for Emergency Responders: During on-site rescue operations, emergency personnel and relevant individuals should, based on the unique hazards present, wear appropriate protective gear as per regulatory requirements, and take suitable safety and protective measures before entering the rescue site. For instance, they should don safety helmets, gloves, protective goggles, and other protective equipment, and wear life jackets, etc.

Rescue of Endangered Individuals: Rescue personnel should carry life-saving equipment and quickly enter the site to move endangered and trapped individuals to safe zones. Those within the caution and isolation zones, who are not involved in the accident's emergency response, should be evacuated to safe zones, choosing the correct direction and route during evacuation. After providing on-site first aid and registration to the rescued individuals, they should be handed over to professional medical and health institutions for further rescue.

Evacuation and Rescue of Personnel: Depending on the accident's situation, the on-site command post should promptly make evacuation decisions and organize the evacuation and rescue of personnel. For example, isolating the leakage pollution zone, restricting personnel access, forming evacuation teams, entering the dangerous leakage zone, and organizing the masses to evacuate along designated routes in the upwind or side upwind direction. Life-saving teams should be formed, equipped with life-saving equipment, they should quickly enter the hazardous area and move all endangered individuals to safe zones.

    \item \textbf{ChatGLM-6b:} Immediate Notification: Upon detecting a hazardous chemical leak, emergency responders should immediately dial the local hazardous chemical leak rescue hotline or contact the local emergency department to report the situation, allowing for appropriate measures to be taken.

Scene Preservation: Responders should strive to minimize damage to the scene, preventing further contamination from the leaked hazardous chemicals. Protective gear such as suits, gloves, masks, etc., should be donned by the responders to avoid contact with the hazardous chemicals and prevent potential infection or poisoning.

Leak Containment: Responders should aim to collect the leaked hazardous chemicals as much as possible, and transfer them to a safe area. Protective gear, such as gloves and masks, can be utilized to collect the hazardous chemicals into plastic bags or containers, preventing direct contact or inhalation.

Rescue Measures Implementation: According to the nature and hazards of the leaked chemicals, suitable rescue measures should be taken, such as isolating the leak area, controlling the leak source, and eliminating the leaked substances. Responders should follow safe operating procedures to avoid causing further danger through inappropriate actions.

Professional Consultation: Responders should communicate with professional hazardous chemical leak rescue personnel to obtain more detailed rescue information and advice. The responders should proceed with the rescue measures according to the professionals' instructions.

Safe Evacuation: Upon completion of the rescue, responders should safely evacuate the scene and return home to rest. During the evacuation process, responders should avoid leaving home or entering hazardous zones to ensure their own safety.

    \item \textbf{GPT-3.5:} Accidents involving hazardous chemical spills require immediate yet cautious response to ensure the safety of the responders and minimize environmental and public impact. The following steps may be employed by emergency responders in such incidents:

Incident Assessment: Initially, a specialized team should be rapidly deployed by the command center to the site for an assessment. This includes determining the nature, quantity, spread, potential harm, and risks of the leaked chemicals. This will aid in developing an appropriate response strategy.

Establishment of Isolation Zones: Based on the assessment, hazard and safe zones should be delineated with a cordon to prevent further human and incident expansion. Unauthorized personnel should not be allowed into the isolated areas.

Selection of Proper Protection Measures: Responders must wear suitable personal protective equipment (PPE) depending on the nature and risks of the leaked chemicals, including protective suits, respirators, goggles, gloves, etc. Correct selection of PPE can reduce the risk of exposure to hazardous materials.

Development of Response Plan: The command center should formulate a response plan based on the incident assessment. This may include choosing the right response tactics, such as leak stoppage, dilution, solidification, etc., to reduce the harm of the chemicals.

Emergency Actions: As per the response plan, emergency actions should be carried out by the responders, such as plugging the leak, removing the spilled substances, diluting concentrations, etc.

Monitoring and Evaluation: Throughout the response, air quality, environmental parameters, and chemical concentrations should be monitored. This can help adjust the response strategy promptly to protect the safety of the responders and the public.

Evacuation: If necessary, nearby residents and personnel should be evacuated. The evacuation process should be orderly to avoid panic and confusion.

Post-Incident Handling: After the incident, post-incident tasks should be performed, including site cleanup, wastewater discharge, and waste handling, to minimize environmental impact.

Incident Investigation and Review: After the response, an investigation should be conducted to analyze the causes of the incident, review lessons learned, and prevent similar future incidents.

Emergency responders must undergo professional training, understanding the nature of hazardous chemicals and response measures, strictly following operating procedures to ensure safety and effectiveness during the response. Collaborations with other relevant departments and organizations are also essential to ensure coordination and efficiency throughout the response process.
    \end{itemize}
\item \textbf{Q3: (Accident Investigation)}A hazardous chemical warehouse has experienced a leakage. Upon inspection, the source of the gas leak was identified as a pipeline leakage. After investigation, natural disasters and human factors have been ruled out. What could be the potential causes of this incident?
    \begin{itemize} 
    \item \textbf{E-KELL:} Potential accident causes include:

Pipeline Quality Defects: Pipeline corrosion is a main cause of pipeline leakage, hence, pipeline quality defects may lead to pipeline corrosion.

Gas Defects: Hazardous material warehouses often need to store various gases. If gas storage is improper or leakage occurs, it may result in pipeline corrosion.

Corrosion Protection Defects: Defects in corrosion protection may also lead to pipeline corrosion as external corrosive substances might damage the pipeline's protective layer.

Pipeline Protection Defects: Pipeline protection defects may cause the pipeline to undergo stress damage, especially during long-term storage and transportation of hazardous materials.

Design Defects: Design defects in pipelines may result in pipe rupture or leakage, particularly at pipe joints or bends.

Stress Prevention Defects: While stress prevention defects may not completely prevent pipeline leakage, they can reduce the stress damage suffered by the pipeline.

Construction Process Defects: Defects during the pipeline construction process could lead to protection or design defects in the pipeline, resulting in pipeline leakage.

Safety Management Defects: Safety management defects may result in improper operation by workers or lack of training, thus leading to pipeline leakage.

Installation Defects: Pipeline installation defects may lead to poor sealing at pipe joints or inherent issues with the pipeline itself, leading to pipeline leakage.
    \item \textbf{ChatGLM-6b:} We can infer the following probable causes for the incident:

Pipeline rupture or damage

Defects in pipeline design or manufacture

Improper pipeline maintenance or upkeep

Poor sealing at pipeline joints

Accidental collisions or compressions during transportation or handling

Other potential causes could include pipeline rupture or damage due to environmental factors such as temperature, pressure, humidity, etc.
    \item \textbf{GPT-3.5:} Upon excluding natural disasters and human factors, potential causes of hazardous materials warehouse pipeline leakage may include:

Equipment failure or aging: Corrosion, wear and tear, or fatigue may occur in pipelines over extended periods of use, leading to leakage. Equipment aging, lack of maintenance, or improper repair could be the root cause of the leak.

Design flaws: If there are deficiencies in the design of the pipeline system, such as issues with material selection, connection methods, or structural design, leakage might occur. 

Construction quality issues: Improper construction, welding, or installation processes could introduce defects, ultimately causing pipeline leakage.

Excessive pressure: Overly high pressure could exceed the pipeline's capacity, triggering explosions or leaks.

Temperature effects: Extreme temperature changes could cause pipeline expansion or contraction, thereby inducing loosening or rupture at connection points.

Chemical reactions: Hazardous materials stored may undergo uncontrolled chemical reactions within the pipeline, resulting in pipeline damage and leakage.

Operational errors: Mistakes made by operators, such as improper operation or incorrect valve operation, could lead to pipeline system rupture.

Insufficient monitoring: Faults or deficiencies in the monitoring system could lead to undetected pipeline leaks, thus delaying emergency responses.

Improper material selection: Choosing materials unsuitable for storing specific hazardous materials could lead to corrosion or erosion of the pipeline, ultimately causing leakage.

Vibration and impact: Shaking or impacting the pipeline system, especially during the transportation of hazardous materials, could damage the pipeline and trigger leakage.

Conducting thorough investigations and analyses of these potential causes is crucial in order to devise effective preventive measures and prevent similar incidents from reoccurring.
    \end{itemize}
\item \textbf{Q4: (Hazard Source Identification):}A leakage incident has transpired at a hazardous chemical warehouse involving the emission of sulfur dioxide and propylene. What are the physical and chemical properties of these substances, and do they pose risks of combustion or explosion?
    \begin{itemize} 
    \item \textbf{E-KELL:} The physicochemical characteristics of sulfur dioxide are a boiling point of -10°C, a gaseous relative density of 2.25, non-flammability, but it can explode upon contact with potassium chlorate at room temperature. Meanwhile, propylene exhibits a gaseous relative density of 1.5 and an explosive limit of 1.0\% to 15.0\%. Therefore, in the event of a leak in a hazardous materials warehouse, where the leaking gases include sulfur dioxide and propylene, both possess explosive hazards, with sulfur dioxide presenting a higher risk of explosion.
    \item \textbf{ChatGLM-6b:} The physicochemical properties of sulfur dioxide (SO2) and propylene (C3H8O) are as follows:

Sulfur dioxide (SO2) is a colorless, toxic, and volatile gas with a molecular formula of SO2 and a molecular weight of approximately 38.94g/mol. Sulfur dioxide possesses strong reducing properties, enabling it to react with oxidants to produce sulfuric acid (H2SO4). It can react with oxygen to form sulfates and water, making it a vital industrial raw material.

Propylene (C3H8O) is a colorless, transparent, sweet-tasting gas with a molecular formula of C3H8O and a molecular weight of approximately 58.05g/mol. Propylene exhibits a high refractive index, high specific surface area, and strong adsorption properties, making it suitable for the preparation of polymers, coatings, plastics, and other materials.

Due to the strong reducing and toxic properties of sulfur dioxide and propylene, as well as their volatility and diffusibility, they present hazards associated with combustion and explosion. Within hazardous goods warehouses, sulfur dioxide and propylene may react with other chemical substances, producing toxic gases and leading to personnel poisoning and property damage. Therefore, effective safety measures must be taken to prevent leaks and combustion or explosion accidents involving sulfur dioxide and propylene.

    \item \textbf{GPT-3.5:} Sulfur dioxide (SO2) and propene (C3H4) are two common chemical substances, each with distinct physicochemical characteristics and safety profiles. Let me elaborate in detail:

Sulfur Dioxide (SO2):

Physicochemical Characteristics: Sulfur dioxide is a colorless gas with an irritating odor, often found in gases produced by combustion or burning of sulfur-containing compounds. At room temperature, it is a water-soluble gas that can form a solution of sulfur dioxide. When dissolved in water, it forms sulfurous acid (H2SO3), rendering the solution acidic.

Hazards: Sulfur dioxide, with its irritating odor, may cause irritation to the respiratory tract and eyes upon high concentration exposure. It is also an asphyxiant gas that can have severe effects on the human respiratory system at high concentrations. Additionally, sulfur dioxide may react chemically with other substances, producing toxic or irritating products.

Propene (C3H4):

Physicochemical Characteristics: Propene is a colorless, flammable gas with a distinctive smell. It is a simple representative of the alkene hydrocarbons, commonly used in the polymer industry for producing plastics, and can also be used in welding and metal cutting.

Hazards: Propene is a flammable gas that may combust when mixed with oxygen in the air under appropriate conditions. High concentrations of propene gas may explode upon encountering a source of ignition or high temperatures. Therefore, it is necessary to take appropriate safety measures when handling propene to prevent its leakage, accumulation, and ignition.

Based on the above information, in the event of a gas leak in a hazardous chemical warehouse containing sulfur dioxide and propene, immediate actions should be taken to ensure workplace safety. Sulfur dioxide could irritate and harm the human respiratory system, while propene poses potential flammability and explosiveness. In handling such situations, comply with relevant regulations and safety measures for hazardous materials handling, ensuring the safety of both employees and the environment.

    \end{itemize}
\item \textbf{Q5: (Fire Extinguishment)}
In case of a fire in chemical warehouse leakage with sulfur dioxide and acetylene gases, how should I conduct firefighting, and what are the key points to consider?

    \begin{itemize} 
    \item \textbf{E-KELL:} In the event of a hazardous materials warehouse leak accompanied by a fire, it is necessary, following safety principles, to remove containers from the fire scene and use ample amounts of water to cool the containers until the fire is extinguished. Additionally, care must be taken not to introduce water into the containers or use other materials to quell the fire to avoid escalating the hazard. Furthermore, specific warnings should be issued regarding sulfur dioxide and propylene, which are toxic and can severely irritate the eyes and respiratory tract. Therefore, during the firefighting process, it is essential that skilled professionals handle the situation to ensure personnel safety.

    \item \textbf{ChatGLM-6b:} In such circumstances, the following steps should be taken to extinguish the fire:

Immediate Reporting: Upon detecting a leakage and consequent fire in a hazardous materials warehouse, it is crucial to report the incident promptly and alert the relevant authorities for intervention.

Ventilation Shut-off: During the fire, ventilation may persist, leading to an increase in toxic gas concentration. Therefore, it is imperative to close ventilation points immediately to prevent the spread of toxic gases.

Fire Extinguisher Usage: If the hazardous materials warehouse contains toxic gases like sulphur dioxide and propylene, fire extinguishers should be employed. When using fire extinguishers, instructions on the apparatus should be strictly followed to ensure correct usage.

Leakage Area Isolation: During the fire extinguishing process, the leakage area should be isolated to prevent the spread of toxic gases to other areas and mitigate potential harm to people.

Site Protection: After extinguishing the fire, the site should be cleared and preserved from further damage. If the hazardous materials warehouse is indoors, it is essential to ensure site safety to prevent accidental contact with residual toxic gases.

Throughout the fire extinguishing process, safety principles should be adhered to, safeguarding oneself and others. Simultaneously, appropriate measures should be taken based on the on-site conditions to minimize injury and loss to the greatest extent.

    \item \textbf{GPT-3.5:} Leaks and fires in hazardous chemical warehouses represent extremely dangerous scenarios that require cautious handling. Below are some key suggestions and points for firefighting under such circumstances:

Ensure Safe Evacuation: Initially, ensure the safe evacuation of all personnel from the hazardous area, staying as far away as possible from the direction of the leaked substance's wind. Make sure individuals wait at safe locations for rescue and instructions.

Alarm and Notification: Immediately dial emergency numbers to alert the fire department, environmental protection agencies, and other relevant bodies. Provide detailed location information, leaked substance details, and fire conditions.

Isolate the Fire Source: Avoid direct contact with the fire source and the leak point to prevent triggering further explosions or fires.

Choice of Fire Extinguishing Agent: Choose the appropriate extinguishing agent based on the nature of the leaked substance. For sulfur dioxide (SO2), a carbon dioxide (CO2) fire extinguisher or dry powder fire extinguisher could be used. For propylene, a foam fire extinguisher may be more suitable due to its flammability.

Wind Direction Consideration: Consider the wind direction when selecting firefighting methods to ensure that the extinguishing agent is not blown towards crowds or other hazardous areas.

Avoid Water: Do not use water to extinguish fires involving propylene, as it reacts with water producing intense combustion. Similarly, water may exacerbate fires or spread leaks with certain chemicals.

Wear Appropriate Equipment: If it is necessary to enter the leak area for firefighting, ensure the wearing of appropriate protective equipment such as chemical protective suits, gas masks, etc.

Stay Away from Toxic Gases: Both propylene and sulfur dioxide are toxic gases harmful to human health. Ensure people stay away from the leaked gases and set up warning lines at the wind direction to prevent people from entering the hazardous area.

Leave it to Professionals: It is best to have professional firefighters or hazardous material handling experts handle hazardous chemical leaks and fires. They possess the required knowledge and experience to handle dangerous situations more safely.

Prevent Leak Expansion: While extinguishing the fire, strive to control the spread of the leak to minimize the escalation of the situation.

In conclusion, safety is always the paramount consideration when dealing with hazardous chemical leaks and fires. Follow the advice of professionals, take appropriate measures, and ensure timely notification of relevant departments to minimize risk and damage as much as possible.
    \end{itemize}
\item \textbf{Q6: (Personal Protective Equipment)}In the event of a leak at a hazardous materials warehouse, with leaked gases comprising sulfur dioxide and propylene and the area being designated as a moderate toxicity, low hazard zone, how should one proceed with individual protection?
    \begin{itemize} 
    \item \textbf{E-KELL:} Both sulfur dioxide and propylene are hazardous materials, and the site is a moderate toxicity, low hazard zone, so individual protection measures must be taken. 
    
For sulfur dioxide, as it's a toxic gas, full-enclosure chemical protective suits are needed to protect the respiratory system and skin. 

For propylene, as it's extremely flammable, simple gas masks and chemical-resistant gloves are necessary to prevent inhalation of harmful gases and contact with fire. While taking individual protective measures, local protection level standards must also be followed, and items like positive pressure air respirators or positive pressure oxygen respirators, as well as chemical-resistant gloves and boots that meet safety regulations must be chosen. 

Lastly, it's important to emphasize that in the event of a leak in a hazardous materials warehouse, immediate action like cutting off the leaking gas source and closing ventilation ports is necessary to avoid poisoning and other dangers. At the same time, the local authorities must be promptly notified so they can take necessary rescue measures and ensure safety.

    \item \textbf{ChatGLM-6b:} Under such circumstances, the following measures should be adopted for personal protection:

Wearing Protective Gear: In the event of a hazardous chemicals warehouse leakage, protective clothing should be worn to shield oneself from the harmful effects of sulfur dioxide and propylene. The protective attire should encompass masks, gloves, goggles, and shoes.

Avoiding Inhalation of Harmful Gases: Sulfur dioxide and propylene are highly irritant, and their inhalation should be avoided as much as possible. At the leakage site, distance yourself from the leakage source and refrain from deep breathing.

Maintaining Ventilation: In case of a hazardous chemicals warehouse leakage, ventilation should be maintained. Open windows or doors to allow air circulation. 

Avoiding Contact with Leaked Substances: At the leakage site, direct contact with the leaked substances should be avoided as much as possible. If contact is unavoidable, gloves and goggles should be worn, and distance from the source of leakage should be maintained.

Maintaining Good Hygiene Practices: At the leakage site, good hygiene practices should be maintained. Hands should be washed frequently and contact with mucous membranes such as eyes, nose, and mouth should be avoided.

Prompt Reporting: In case of a leakage, authorities should be alerted promptly. Timely reporting can facilitate immediate response, thus preventing further harm.
    \item \textbf{GPT-3.5:} In the event of a hazardous material warehouse leakage, ensuring personal safety is paramount. Given the presence of sulphur dioxide and propylene in the gas leak, and the location being a mild danger zone for poisoning, the following are individual protective measures you can undertake:

Don appropriate Personal Protective Equipment (PPE):

Wear a chemical protective suit to block harmful gases.
Use chemical goggles or face shields to protect your eyes from harmful gas irritants.
Don chemical resistant gloves to avoid direct contact with harmful substances.
Use Respiratory Protective Equipment (RPE):

Considering the gas leakage, ensure you wear appropriate respiratory protective equipment. For sulphur dioxide and propylene, a suitable gas filter respirator or supplied air respirator is required.
Avoid entering the danger zone:

Avoid entering the leakage area unless necessary. Do not attempt to handle the leak if untrained.
Maintain communication with emergency personnel:

Notify the emergency response team as soon as possible so that they can take appropriate measures. Maintain contact with the rescuers while awaiting rescue.
Immediate evacuation:

In case of severe gas leakage, evacuate the danger zone immediately and move to a safe area. Do not take unnecessary risks.
Personal hygiene and cleansing:

If the harmful gas has come in contact with your skin, wash the affected area with plenty of water as soon as possible and change into clean clothes.
Medical evaluation:

Seek medical evaluation as soon as possible, even without obvious symptoms, to ensure you have not been affected by the harmful gas.
Follow emergency plans:

There should be an emergency plan for chemical leaks in the workplace. Follow the steps outlined in the plan.
Note that the above is general guidance, and specific circumstances may differ. The best practice in dealing with chemical leaks is to rely on professional emergency responders and plans. In dangerous situations, protecting your life is the most important thing.

    \end{itemize}
\item \textbf{Q7: (Rescuing)}In the event of a chemical leak in a hazardous materials warehouse, where the leaking gas contains sulfur dioxide and propylene, and there are individuals on-site who have inhaled this gas, how should I proceed with immediate first aid measures?
    \begin{itemize} 
    \item \textbf{E-KELL:} In this scenario, the following immediate emergency measures should be taken:

Quickly evacuate from the scene to a location with fresh air, ensuring the airway remains unobstructed.

In the case of sulfur dioxide inhalation, immediately remove contaminated clothing and thoroughly rinse the skin and eyes with copious amounts of flowing clean water.

In the case of propylene inhalation, seek medical attention immediately and inform the doctor of the type of gas inhaled.

While providing on-site first aid, notify local emergency response personnel and evacuate according to their instructions.

Avoid using hot water or radiant heat to treat the skin of trapped individuals to prevent frostbite.

Administer on-site first aid to rescued individuals and record their information, then transfer them to professional medical and health institutions for further treatment.

Both sulfur dioxide and propylene are toxic gases with strong irritant effects on the respiratory tract and eyes. Therefore, extra caution should be exercised during emergency aid to avoid inadvertent harm to others. Simultaneously, adhere to the instructions of local emergency response personnel to ensure scene safety.
    \item \textbf{ChatGLM-6b:} In the event of a leakage at a hazardous chemical warehouse, with the leaked gases containing sulfur dioxide and propylene, the on-site staff should immediately implement the following emergency response measures:

Immediate Ventilation Shutdown: Both sulfur dioxide and propylene are toxic gases. Shutting down the ventilation system can reduce the concentration of these substances inhaled, thus lessening harm to the human body.

Personnel Evacuation: On-site staff should immediately evacuate to a safe zone to avoid inhaling toxic gases.

Immediate Ventilation Upon Inhalation: If individuals have inhaled sulfur dioxide and propylene, immediate ventilation should be provided to expel the gases from the body.

Respiratory Equipment Usage: If individuals need to breathe fresh air, respiratory equipment should be worn to maintain respiration.

Contact Emergency Personnel: If on-site staff feel unwell, they should immediately contact emergency personnel for intervention.

Before undertaking emergency response measures, staff should stay as far away from the leakage area as possible to avoid gas inhalation. If individuals feel unwell, immediate medical attention should be sought.
    \item \textbf{GPT-3.5:} In the event of a leakage in a hazardous chemical warehouse, with individuals inhaling released gases (sulphur dioxide and propylene), prompt emergency response measures must be taken. The following are some emergency first aid steps:

Ensure your safety: Prior to initiating any form of aid, ensure your own safety. If possible, don appropriate protective gear to avoid exposure to the toxic gases.

Move victims from the danger area: Swiftly relocate individuals who have inhaled the gases from the leakage zone to areas with good air circulation. If the toxic gas leakage continues, ensure the evacuation direction is not downwind.

Ventilation and fresh air: Place the victims outdoors, ensuring access to ample fresh air. Open windows and doors to facilitate proper ventilation.

Call for emergency assistance: Simultaneously with providing first aid, immediately dial the local emergency number, alerting medical professionals and the fire department. Provide detailed location information and symptoms of the victims.

Check breathing and consciousness: Verify if the victims are still breathing. If a victim is not breathing, start Cardiopulmonary Resuscitation (CPR) immediately.

Provide oxygen: If oxygen equipment is available, try to administer oxygen to the victims. This can help the victims recover from toxic gas exposure more quickly.

Monitor symptoms: Whilst waiting for medical assistance to arrive, continually monitor the victims' symptoms. If symptoms worsen, ensure medical aid handles it promptly.

Do not delay: Take first aid measures without delay, as exposure to toxic gases can lead to serious health problems. Your first aid could be crucial in saving lives before professional rescuers arrive.

Please note, dealing with toxic gas exposure requires the involvement of medical professionals. The above first aid steps are interim measures; ensure the best care for the victims until medical professionals arrive. Also, to prevent similar incidents from recurring, handling and storage of hazardous chemicals should strictly adhere to relevant laws and safety standards.
    \end{itemize}
\item \textbf{Q8: (Isolation)}In the event of chemical leakage from a hazardous materials warehouse with sulfur dioxide and propylene gases, how should I isolate and evacuate the nearby population to ensure public safety?
    \begin{itemize} 
    \item \textbf{E-KELL:} In the event of a hazardous materials warehouse leak, the following measures should be taken to ensure public safety:

Initial isolation measures should be taken to prevent the spread of contamination to surrounding areas and limit the access of individuals when a hazardous materials warehouse leak occurs. If the extent of contamination is unclear, an initial isolation of at least 500 meters should be imposed, with downwind evacuation extending at least 1500 meters. Subsequently, gas concentration tests should be performed, and based on the actual concentration of harmful gases, the isolation and evacuation distances should be adjusted.

An evacuation team should be formed to enter the leak hazard area and organize the mass evacuation along designated routes in the upwind or side upwind direction. 

Should the hazardous materials warehouse contain storage tanks, tank trucks, or tank cars, an isolation of 1600 meters should be imposed. 

Rescued individuals should be registered, identified, and provided with on-site aid. 

Those requiring medical treatment should be handed over to medical emergency departments for care.
    \item \textbf{ChatGLM-6b:} Under such circumstances, the following immediate actions should be taken:

Isolate the Leak Zone: Separate the leakage area from the surrounding regions to prevent the expansion of contamination.

Evacuate Personnel: Notify residents in proximity to evacuate to avoid contact with the leaked chemicals.

Contact Professionals: Reach out to the local emergency service teams, providing them with detailed information, enabling them to take appropriate measures.

Implement Safety Measures: After personnel evacuation, undertake necessary safety precautions such as wearing masks and gloves to prevent contamination spread.

Monitor Contamination: Closely monitor the state of contamination, ensuring timely measures are taken to address it.

During the process of isolation and evacuation of personnel, it is crucial to adhere to the guidance of local emergency service teams and provide as much detailed information as possible for them to take appropriate actions.
    \item \textbf{GPT-3.5:} A hazardous material leak constitutes a serious situation, with the safety of nearby residents being of paramount importance. Here are some steps that should be taken to isolate and evacuate the local population to ensure public safety:

Immediate Alert: Notify local emergency services, fire departments, and environmental protection agencies, providing information about the nature of the hazardous substances and the extent of the leak so they can respond promptly.

Identify Leak Source and Spread Direction: Ascertain the location and origin of the leak as well as the direction of air dispersion, which will help determine the scope of the safe zone.

Establish Isolation Zone: Based on the source of the leak and the direction of air dispersion, create an isolation zone to prevent any personnel from entering the danger zone. This can be marked with caution tape, barriers, etc.

Initiate Evacuation Plan: Develop an evacuation plan based on the leak status and wind direction, and guide nearby residents, workers, and students to safety according to this plan.

Provide Protective Measures: Provide necessary protective gear such as respirators and protective suits to those who may potentially be affected.

Information Dissemination and Publicity: Use broadcast media, text messages, social media, etc., to disseminate leakage information, evacuation instructions, and safety advice to nearby residents.

Coordinate Rescue Efforts: Collaborate with emergency services and professional rescue teams in cleaning up and managing the leaked substances to ensure the leak is controlled.

Monitoring and Assessment: After controlling the leak, monitor the site to ascertain that the danger has been eliminated. Subsequently, assess the situation, checking for any potential safety hazards.

Cease Alarm: Once the danger has been controlled and the safe zone restored to normal, promptly discontinue the alarm and notify relevant personnel.

Post-incident Management: Organize post-incident management and accident investigation after the event, analyze the cause of the accident, and draw lessons to improve response capabilities for similar incidents in the future.

Please note, handling hazardous material leaks requires the involvement of professionals, and these steps are for reference only. In any emergency situation, please follow the directives and advice of your local emergency services.

    \end{itemize}
\item \textbf{Q9: (Hazard Source Management)}In the event of a leakage in a hazardous chemical warehouse, where the leaking gases include sulfur dioxide and propylene, what would be the appropriate procedure for managing the leakage?
    \begin{itemize} 
    \item \textbf{E-KELL:} For instances of leakage in hazardous chemical warehouses, the following leak control measures should be adopted:

Upon ensuring safety, measures like closing valves and sealing leaks should be implemented to cut off the leakage source. It is crucial to prevent the gas from spreading through sewer systems, ventilation systems, or entering confined spaces. Water in a spray form should be used to dissolve and dilute the leaking gas. The leakage area should be isolated until the gas completely dissipates.

For handling leaks of sulfur dioxide and propylene, all ignition sources should be eliminated (smoking is prohibited near the leakage area, all open flames, sparks or flames should be extinguished). Explosion-proof communication tools should be used, and all equipment should be grounded during operations. Measures like closing valves and sealing leaks should be used to cut off the leakage source, upon ensuring safety. It is important to prevent the gas from spreading through sewer systems, ventilation systems, or entering confined spaces. Water in a spray form should be used to change the direction of the vapor cloud. The leakage area should be isolated until the gas completely dissipates. Moreover, it is necessary to specifically warn that sulfur dioxide is toxic and has a strong irritating effect on the eyes and respiratory tract mucosa. Extinguishing flames at the leakage site is strictly prohibited.
    \item \textbf{ChatGLM-6b:} Under these circumstances, the following actions should be immediately taken:

Shut down the leakage source promptly. This can be achieved by closing valves, ventilation ducts, or doors.

Evacuate personnel. If the area is densely populated, individuals should be immediately evacuated to safe zones.

Collect the leaking material. It should be promptly gathered and sent for processing to local environmental protection agencies or relevant departments.

Contact local environmental protection agencies or relevant departments. They should be immediately informed about the leakage incident.

Take necessary measures to prevent further leakage. This might include the use of gas control devices, increased ventilation, and other actions to lower the concentration of the leaking substance.

Await professional handling. Necessary measures should continue to be taken to ensure the leakage is under control, until professional help arrives.
    \item \textbf{GPT-3.5:} Addressing hazardous chemical leaks is a risky and complex task that requires specialized knowledge and equipment. For leaks of sulfur dioxide and propylene, here are some potential response steps:

Personal Protective Equipment (PPE): Before addressing the leak, ensure proper PPE is worn, including chemical protective clothing, gas masks, goggles, gloves, and slip-resistant shoes.

Isolate the Leak Source: As much as possible, isolate the leak source to prevent further spread. This may involve shutting off valves, powering down equipment, or sealing off the leaking pipe or container.

Wind Direction Determination: Identify the wind direction to decide where to establish a safety zone, evacuation route, and work direction.

Emergency Response: For smaller-scale leaks, consider using appropriate chemical absorbents or adsorbents to soak up the leaked material. Ensure that the adsorbents used are compatible with the leaked substance.

Leak Spread Control: Utilize equipment such as windbreak walls or water dams to limit the spread of the leak.

Liquid Leak Management: For liquid leaks, use appropriate chemical absorbents for adsorption. Collect contaminated absorbents and leaked substances for proper disposal.

Gas Leak Management: For gas leaks, the top priority is to protect personnel safety. In cases of toxic gas leaks, avoid proximity to the leak source and quickly evacuate to a safe location upwind.

Professional Team Intervention: For large-scale or complex leaks, immediately contact a professional hazardous materials response team. They have the necessary expertise and equipment to handle hazardous material leaks.

Ventilation and Cleanup: Once the leak is under control, ensure good ventilation conditions while cleaning up the site to prevent residual hazardous substances from polluting the environment.

Post-incident Management: After the leak incident is handled, conduct post-incident management and accident investigation to learn from the experience and improve future emergency response capabilities.

Please note, managing hazardous chemical leaks requires adherence to relevant laws and safety standards to ensure the process does not cause further harm to personnel, the environment, or property. Under all circumstances, cooperate with professional emergency departments or hazardous material handling experts and follow their advice.
    \end{itemize}
\item \textbf{Q10: (Report Compilation)}What information should be included in the compilation of firefighting?
    \begin{itemize} 
    \item \textbf{E-KELL:} Writing about the firefighting situation should include the following information:

Firefighting Actions Taken: This can include the use of fire extinguishers, calling the fire department, dispatching firefighters, and so on.

Utilization of Firefighting Water Sources: This can encompass the quantity, location, and type of firefighting water sources.

Control of Fire Duration: This can include the start time, end time, and ongoing timeline of fire control.

Extinguishing Time: This can include the start time, end time, and ongoing timeline of extinguishing the fire.

Area of Fire Spread: This can involve the size and location of the fire spread.
Number of People Evacuated and Rescued by Firefighters: This can include the number of people evacuated and rescued by firefighters.

Value of Protected Property: This can entail the value of the protected property and the measures taken to safeguard it.
. 
    \item \textbf{ChatGLM-6b:} When documenting fire fighting situations, the following information should be included:

The time and location of the fire fight

The objectives of the fire fight

The methods and steps undertaken in the fire fight

The results and impacts of the fire fight

Other relevant information, such as casualties and property damage.
    \item \textbf{GPT-3.5:} The report on the firefighting situation should include the following key information:

Time and Location: Indicate the exact time and location of the firefighting situation, including the geographical coordinates and specific description of the incident site.

Cause of Fire: Describe the cause of the fire, which may be natural factors (such as lightning, drought, etc.) or human factors (such as ignition source, point of fire, etc.).

Fire Spread: Detailed description of the fire expansion, including the direction, speed, and scope of fire spread.

Affected Areas: Explain the impact of the fire on the surrounding areas, including affected residents, buildings, wildlife, natural resources, etc.

Casualties and Evacuation: Report casualties caused by the fire, including firefighters and residents, as well as the progress of evacuation efforts.

Firefighting Actions: Describe the firefighting actions taken by the fire department, including manpower, resources, equipment used, and specific measures taken, such as establishing firebreaks, using fire extinguishing agents, etc.

Assisting Forces: Mention other organizations or forces involved in firefighting, such as national rescue teams, the military, and their coordination and support.

Weather Conditions: Analyze the weather conditions at the time of the fire, including temperature, humidity, wind direction and speed, and other meteorological factors affecting the fire spread.

Protective Measures: Report protective measures taken by residents and rescue workers, such as wearing masks, evacuation guidelines, etc., to ensure the safety of personnel.

Firefighting Progress: Update the progress of firefighting actions, including the length of fire lines controlled, fire sources extinguished, etc.

Difficulties and Challenges: Describe the difficulties and challenges encountered in the firefighting process, such as harsh weather conditions, complex terrain, etc.

Media and Public Relations: Document the communication and cooperation between government departments and the media and public, including information dissemination, provision of guides, etc.

Social Support: Mention the support provided by volunteers, community organizations, and the assistance and help provided by people for the victims.

Post-fire Assessment: After the firefighting is over, carry out an assessment of the fire impact, including an analysis of damaged buildings, environmental impact, etc.

Future Prevention: Discuss future prevention measures for similar fires, such as strengthening forest management, promoting public fire awareness, etc.

This information will help to present the firefighting situation in a comprehensive and objective manner, providing the public with accurate information and understanding.
    \end{itemize}
\end{enumerate}  \textbf{}

\printcredits

\bibliographystyle{cas-model2-names}

\bibliography{cas-refs}

\begin{thebibliography}{37}
\expandafter\ifx\csname natexlab\endcsname\relax\def\natexlab#1{#1}\fi
\providecommand{\url}[1]{\texttt{#1}}
\providecommand{\href}[2]{#2}
\providecommand{\path}[1]{#1}
\providecommand{\DOIprefix}{doi:}
\providecommand{\ArXivprefix}{arXiv:}
\providecommand{\URLprefix}{URL: }
\providecommand{\Pubmedprefix}{pmid:}
\providecommand{\doi}[1]{\href{http://dx.doi.org/#1}{\path{#1}}}
\providecommand{\Pubmed}[1]{\href{pmid:#1}{\path{#1}}}
\providecommand{\bibinfo}[2]{#2}
\ifx\xfnm\relax \def\xfnm[#1]{\unskip,\space#1}\fi
\bibitem[{Choudhary and Reddy(2023)}]{choudhary2023complex}
\bibinfo{author}{Choudhary, N.}, \bibinfo{author}{Reddy, C.K.},
  \bibinfo{year}{2023}.
\newblock \bibinfo{title}{Complex logical reasoning over knowledge graphs using
  large language models}.
\newblock \bibinfo{journal}{arXiv preprint arXiv:2305.01157} .
\bibitem[{Cremen et~al.(2022)Cremen, Bozzoni, Pistorio and
  Galasso}]{cremen2022developing}
\bibinfo{author}{Cremen, G.}, \bibinfo{author}{Bozzoni, F.},
  \bibinfo{author}{Pistorio, S.}, \bibinfo{author}{Galasso, C.},
  \bibinfo{year}{2022}.
\newblock \bibinfo{title}{Developing a risk-informed decision-support system
  for earthquake early warning at a critical seaport}.
\newblock \bibinfo{journal}{Reliability Engineering \& System Safety}
  \bibinfo{volume}{218}, \bibinfo{pages}{108035}.
\bibitem[{Deveci et~al.(2022)Deveci, Pamucar, Gokasar, Delen and
  Mart{\'\i}nez}]{deveci2022fuzzy}
\bibinfo{author}{Deveci, M.}, \bibinfo{author}{Pamucar, D.},
  \bibinfo{author}{Gokasar, I.}, \bibinfo{author}{Delen, D.},
  \bibinfo{author}{Mart{\'\i}nez, L.}, \bibinfo{year}{2022}.
\newblock \bibinfo{title}{A fuzzy einstein-based decision support system for
  public transportation management at times of pandemic}.
\newblock \bibinfo{journal}{Knowledge-Based Systems} \bibinfo{volume}{252},
  \bibinfo{pages}{109414}.
\bibitem[{Ding et~al.(2023)Ding, Qin, Yang, Wei, Yang, Su, Hu, Chen, Chan, Chen
  et~al.}]{ding2023parameter}
\bibinfo{author}{Ding, N.}, \bibinfo{author}{Qin, Y.}, \bibinfo{author}{Yang,
  G.}, \bibinfo{author}{Wei, F.}, \bibinfo{author}{Yang, Z.},
  \bibinfo{author}{Su, Y.}, \bibinfo{author}{Hu, S.}, \bibinfo{author}{Chen,
  Y.}, \bibinfo{author}{Chan, C.M.}, \bibinfo{author}{Chen, W.}, et~al.,
  \bibinfo{year}{2023}.
\newblock \bibinfo{title}{Parameter-efficient fine-tuning of large-scale
  pre-trained language models}.
\newblock \bibinfo{journal}{Nature Machine Intelligence} \bibinfo{volume}{5},
  \bibinfo{pages}{220--235}.
\bibitem[{Fan et~al.(2023)Fan, Jiang, Li and Li}]{fan2023grammargpt}
\bibinfo{author}{Fan, Y.}, \bibinfo{author}{Jiang, F.}, \bibinfo{author}{Li,
  P.}, \bibinfo{author}{Li, H.}, \bibinfo{year}{2023}.
\newblock \bibinfo{title}{Grammargpt: Exploring open-source llms for native
  chinese grammatical error correction with supervised fine-tuning}, in:
  \bibinfo{booktitle}{CCF International Conference on Natural Language
  Processing and Chinese Computing}, \bibinfo{organization}{Springer}. pp.
  \bibinfo{pages}{69--80}.
\bibitem[{Fertier et~al.(2020)Fertier, Barthe-Delano{\"e}, Montarnal, Truptil
  and B{\'e}naben}]{fertier2020new}
\bibinfo{author}{Fertier, A.}, \bibinfo{author}{Barthe-Delano{\"e}, A.M.},
  \bibinfo{author}{Montarnal, A.}, \bibinfo{author}{Truptil, S.},
  \bibinfo{author}{B{\'e}naben, F.}, \bibinfo{year}{2020}.
\newblock \bibinfo{title}{A new emergency decision support system: The
  automatic interpretation and contextualisation of events to model a crisis
  situation in real-time}.
\newblock \bibinfo{journal}{Decision Support Systems} \bibinfo{volume}{133},
  \bibinfo{pages}{113260}.
\bibitem[{Goecks and Waytowich(2023)}]{goecks2023disasterresponsegpt}
\bibinfo{author}{Goecks, V.G.}, \bibinfo{author}{Waytowich, N.R.},
  \bibinfo{year}{2023}.
\newblock \bibinfo{title}{Disasterresponsegpt: Large language models for
  accelerated plan of action development in disaster response scenarios}.
\newblock \bibinfo{journal}{arXiv preprint arXiv:2306.17271} .
\bibitem[{G{\"u}ler and Ge{\c{c}}ici(2020)}]{guler2020decision}
\bibinfo{author}{G{\"u}ler, M.G.}, \bibinfo{author}{Ge{\c{c}}ici, E.},
  \bibinfo{year}{2020}.
\newblock \bibinfo{title}{A decision support system for scheduling the shifts
  of physicians during covid-19 pandemic}.
\newblock \bibinfo{journal}{Computers \& Industrial Engineering}
  \bibinfo{volume}{150}, \bibinfo{pages}{106874}.
\bibitem[{Head et~al.(2023)Head, Jasper, McConnachie, Raftree and
  Higdon}]{head2023large}
\bibinfo{author}{Head, C.B.}, \bibinfo{author}{Jasper, P.},
  \bibinfo{author}{McConnachie, M.}, \bibinfo{author}{Raftree, L.},
  \bibinfo{author}{Higdon, G.}, \bibinfo{year}{2023}.
\newblock \bibinfo{title}{Large language model applications for evaluation:
  Opportunities and ethical implications}.
\newblock \bibinfo{journal}{New Directions for Evaluation}
  \bibinfo{volume}{2023}, \bibinfo{pages}{33--46}.
\bibitem[{Huang et~al.(2023)Huang, Wei, Yue, Zhao, Chen, Li, Jiang, Chang,
  Zhang, Zhang et~al.}]{huang2023dsqa}
\bibinfo{author}{Huang, D.}, \bibinfo{author}{Wei, Z.}, \bibinfo{author}{Yue,
  A.}, \bibinfo{author}{Zhao, X.}, \bibinfo{author}{Chen, Z.},
  \bibinfo{author}{Li, R.}, \bibinfo{author}{Jiang, K.},
  \bibinfo{author}{Chang, B.}, \bibinfo{author}{Zhang, Q.},
  \bibinfo{author}{Zhang, S.}, et~al., \bibinfo{year}{2023}.
\newblock \bibinfo{title}{Dsqa-llm: Domain-specific intelligent question
  answering based on large language model}, in:
  \bibinfo{booktitle}{International Conference on AI-generated Content},
  \bibinfo{organization}{Springer}. pp. \bibinfo{pages}{170--180}.
\bibitem[{Imran et~al.(2020)Imran, Ofli, Caragea and Torralba}]{imran2020using}
\bibinfo{author}{Imran, M.}, \bibinfo{author}{Ofli, F.},
  \bibinfo{author}{Caragea, D.}, \bibinfo{author}{Torralba, A.},
  \bibinfo{year}{2020}.
\newblock \bibinfo{title}{Using ai and social media multimodal content for
  disaster response and management: Opportunities, challenges, and future
  directions}.
\bibitem[{Jiang and Yang(2023)}]{jiang2023legal}
\bibinfo{author}{Jiang, C.}, \bibinfo{author}{Yang, X.}, \bibinfo{year}{2023}.
\newblock \bibinfo{title}{Legal syllogism prompting: Teaching large language
  models for legal judgment prediction}, in: \bibinfo{booktitle}{Proceedings of
  the Nineteenth International Conference on Artificial Intelligence and Law},
  pp. \bibinfo{pages}{417--421}.
\bibitem[{Jiang et~al.(2023)Jiang, Liu, Nejatian, Nasir-Moin, Wang, Abidin,
  Eaton, Riina, Laufer, Punjabi et~al.}]{jiang2023health}
\bibinfo{author}{Jiang, L.Y.}, \bibinfo{author}{Liu, X.C.},
  \bibinfo{author}{Nejatian, N.P.}, \bibinfo{author}{Nasir-Moin, M.},
  \bibinfo{author}{Wang, D.}, \bibinfo{author}{Abidin, A.},
  \bibinfo{author}{Eaton, K.}, \bibinfo{author}{Riina, H.A.},
  \bibinfo{author}{Laufer, I.}, \bibinfo{author}{Punjabi, P.}, et~al.,
  \bibinfo{year}{2023}.
\newblock \bibinfo{title}{Health system-scale language models are all-purpose
  prediction engines}.
\newblock \bibinfo{journal}{Nature} , \bibinfo{pages}{1--6}.
\bibitem[{Jiang et~al.(2022)Jiang, Zhang and Pian}]{jiang2022chatbot}
\bibinfo{author}{Jiang, Q.}, \bibinfo{author}{Zhang, Y.},
  \bibinfo{author}{Pian, W.}, \bibinfo{year}{2022}.
\newblock \bibinfo{title}{Chatbot as an emergency exist: Mediated empathy for
  resilience via human-ai interaction during the covid-19 pandemic}.
\newblock \bibinfo{journal}{Information processing \& management}
  \bibinfo{volume}{59}, \bibinfo{pages}{103074}.
\bibitem[{Jovanović and Campbell(2023)}]{10286238}
\bibinfo{author}{Jovanović, M.}, \bibinfo{author}{Campbell, M.},
  \bibinfo{year}{2023}.
\newblock \bibinfo{title}{Connecting ai: Merging large language models and
  knowledge graph}.
\newblock \bibinfo{journal}{Computer} \bibinfo{volume}{56},
  \bibinfo{pages}{103--108}.
\bibitem[{Le et~al.(2023)Le, Le and Le}]{le2023knowledge}
\bibinfo{author}{Le, T.}, \bibinfo{author}{Le, N.}, \bibinfo{author}{Le, B.},
  \bibinfo{year}{2023}.
\newblock \bibinfo{title}{Knowledge graph embedding by relational rotation and
  complex convolution for link prediction}.
\newblock \bibinfo{journal}{Expert Systems with Applications}
  \bibinfo{volume}{214}, \bibinfo{pages}{119122}.
\bibitem[{Levy et~al.(2016)Levy, Gopalakrishnan and Lin}]{levy2016advances}
\bibinfo{author}{Levy, J.K.}, \bibinfo{author}{Gopalakrishnan, C.},
  \bibinfo{author}{Lin, Z.}, \bibinfo{year}{2016}.
\newblock \bibinfo{title}{Advances in decision support systems for flood
  disaster management: Challenges and opportunities}.
\newblock \bibinfo{journal}{Water Resources and Decision-Making Systems} ,
  \bibinfo{pages}{81--100}.
\bibitem[{Liga and Robaldo(2023)}]{liga2023fine}
\bibinfo{author}{Liga, D.}, \bibinfo{author}{Robaldo, L.},
  \bibinfo{year}{2023}.
\newblock \bibinfo{title}{Fine-tuning gpt-3 for legal rule classification}.
\newblock \bibinfo{journal}{Computer Law \& Security Review}
  \bibinfo{volume}{51}, \bibinfo{pages}{105864}.
\bibitem[{Mansurova et~al.(2023)Mansurova, Nugumanova and
  Makhambetova}]{mansurova2023development}
\bibinfo{author}{Mansurova, A.}, \bibinfo{author}{Nugumanova, A.},
  \bibinfo{author}{Makhambetova, Z.}, \bibinfo{year}{2023}.
\newblock \bibinfo{title}{Development of a question answering chatbot for
  blockchain domain}.
\newblock \bibinfo{journal}{Scientific Journal of Astana IT University} ,
  \bibinfo{pages}{27--40}.
\bibitem[{Martino et~al.(2023)Martino, Iannelli and
  Truong}]{martino2023knowledge}
\bibinfo{author}{Martino, A.}, \bibinfo{author}{Iannelli, M.},
  \bibinfo{author}{Truong, C.}, \bibinfo{year}{2023}.
\newblock \bibinfo{title}{Knowledge injection to counter large language model
  (llm) hallucination}, in: \bibinfo{booktitle}{European Semantic Web
  Conference}, \bibinfo{organization}{Springer}. pp. \bibinfo{pages}{182--185}.
\bibitem[{Mehta and Kumar(2023)}]{mehta2023enabling}
\bibinfo{author}{Mehta, C.}, \bibinfo{author}{Kumar, K.}, \bibinfo{year}{2023}.
\newblock \bibinfo{title}{Enabling knowledge discovery in natural hazard
  engineering datasets on designsafe}.
\newblock \bibinfo{journal}{arXiv preprint arXiv:2304.11273} .
\bibitem[{Mesk{\'o} and Topol(2023)}]{mesko2023imperative}
\bibinfo{author}{Mesk{\'o}, B.}, \bibinfo{author}{Topol, E.J.},
  \bibinfo{year}{2023}.
\newblock \bibinfo{title}{The imperative for regulatory oversight of large
  language models (or generative ai) in healthcare}.
\newblock \bibinfo{journal}{npj Digital Medicine} \bibinfo{volume}{6},
  \bibinfo{pages}{120}.
\bibitem[{Nussbaumer et~al.(2023)Nussbaumer, Pope and
  Neville}]{nussbaumer2023framework}
\bibinfo{author}{Nussbaumer, A.}, \bibinfo{author}{Pope, A.},
  \bibinfo{author}{Neville, K.}, \bibinfo{year}{2023}.
\newblock \bibinfo{title}{A framework for applying ethics-by-design to decision
  support systems for emergency management}.
\newblock \bibinfo{journal}{Information Systems Journal} \bibinfo{volume}{33},
  \bibinfo{pages}{34--55}.
\bibitem[{Osatuyi and Mendon{\c{c}}a(2013)}]{osatuyi2013temporal}
\bibinfo{author}{Osatuyi, B.}, \bibinfo{author}{Mendon{\c{c}}a, D.},
  \bibinfo{year}{2013}.
\newblock \bibinfo{title}{Temporal modeling of group information foraging: An
  application to emergency response}.
\newblock \bibinfo{journal}{Information processing \& management}
  \bibinfo{volume}{49}, \bibinfo{pages}{169--178}.
\bibitem[{Pan et~al.(2023)Pan, Albalak, Wang and Wang}]{pan2023logic}
\bibinfo{author}{Pan, L.}, \bibinfo{author}{Albalak, A.},
  \bibinfo{author}{Wang, X.}, \bibinfo{author}{Wang, W.Y.},
  \bibinfo{year}{2023}.
\newblock \bibinfo{title}{Logic-lm: Empowering large language models with
  symbolic solvers for faithful logical reasoning}.
\newblock \bibinfo{journal}{arXiv preprint arXiv:2305.12295} .
\bibitem[{Peng et~al.(2023)Peng, Galley, He, Cheng, Xie, Hu, Huang, Liden, Yu,
  Chen et~al.}]{peng2023check}
\bibinfo{author}{Peng, B.}, \bibinfo{author}{Galley, M.}, \bibinfo{author}{He,
  P.}, \bibinfo{author}{Cheng, H.}, \bibinfo{author}{Xie, Y.},
  \bibinfo{author}{Hu, Y.}, \bibinfo{author}{Huang, Q.},
  \bibinfo{author}{Liden, L.}, \bibinfo{author}{Yu, Z.}, \bibinfo{author}{Chen,
  W.}, et~al., \bibinfo{year}{2023}.
\newblock \bibinfo{title}{Check your facts and try again: Improving large
  language models with external knowledge and automated feedback}.
\newblock \bibinfo{journal}{arXiv preprint arXiv:2302.12813} .
\bibitem[{Qi et~al.(2023)Qi, Cao, Rao, Wang, Xiao and Wang}]{qi2023limitation}
\bibinfo{author}{Qi, S.}, \bibinfo{author}{Cao, Z.}, \bibinfo{author}{Rao, J.},
  \bibinfo{author}{Wang, L.}, \bibinfo{author}{Xiao, J.},
  \bibinfo{author}{Wang, X.}, \bibinfo{year}{2023}.
\newblock \bibinfo{title}{What is the limitation of multimodal llms? a deeper
  look into multimodal llms through prompt probing}.
\newblock \bibinfo{journal}{Information Processing \& Management}
  \bibinfo{volume}{60}, \bibinfo{pages}{103510}.
\bibitem[{Ray(2023)}]{RAY2023121}
\bibinfo{author}{Ray, P.P.}, \bibinfo{year}{2023}.
\newblock \bibinfo{title}{Chatgpt: A comprehensive review on background,
  applications, key challenges, bias, ethics, limitations and future scope}.
\newblock \bibinfo{journal}{Internet of Things and Cyber-Physical Systems}
  \bibinfo{volume}{3}, \bibinfo{pages}{121--154}.
\bibitem[{Salvagno et~al.(2023)Salvagno, Taccone and
  Gerli}]{salvagno2023artificial}
\bibinfo{author}{Salvagno, M.}, \bibinfo{author}{Taccone, F.S.},
  \bibinfo{author}{Gerli, A.G.}, \bibinfo{year}{2023}.
\newblock \bibinfo{title}{Artificial intelligence hallucinations}.
\newblock \bibinfo{journal}{Critical Care} \bibinfo{volume}{27},
  \bibinfo{pages}{1--2}.
\bibitem[{Shi et~al.(2023)Shi, Ma, Zhong, Mai, Li, Liu and
  Huang}]{shi2023chatgraph}
\bibinfo{author}{Shi, Y.}, \bibinfo{author}{Ma, H.}, \bibinfo{author}{Zhong,
  W.}, \bibinfo{author}{Mai, G.}, \bibinfo{author}{Li, X.},
  \bibinfo{author}{Liu, T.}, \bibinfo{author}{Huang, J.}, \bibinfo{year}{2023}.
\newblock \bibinfo{title}{Chatgraph: Interpretable text classification by
  converting chatgpt knowledge to graphs}.
\newblock \bibinfo{journal}{arXiv preprint arXiv:2305.03513} .
\bibitem[{Singhal et~al.(2023)Singhal, Azizi, Tu, Mahdavi, Wei, Chung, Scales,
  Tanwani, Cole-Lewis, Pfohl et~al.}]{singhal2023large}
\bibinfo{author}{Singhal, K.}, \bibinfo{author}{Azizi, S.},
  \bibinfo{author}{Tu, T.}, \bibinfo{author}{Mahdavi, S.S.},
  \bibinfo{author}{Wei, J.}, \bibinfo{author}{Chung, H.W.},
  \bibinfo{author}{Scales, N.}, \bibinfo{author}{Tanwani, A.},
  \bibinfo{author}{Cole-Lewis, H.}, \bibinfo{author}{Pfohl, S.}, et~al.,
  \bibinfo{year}{2023}.
\newblock \bibinfo{title}{Large language models encode clinical knowledge}.
\newblock \bibinfo{journal}{Nature} \bibinfo{volume}{620},
  \bibinfo{pages}{172--180}.
\bibitem[{Thirunavukarasu et~al.(2023)Thirunavukarasu, Ting, Elangovan,
  Gutierrez, Tan and Ting}]{thirunavukarasu2023large}
\bibinfo{author}{Thirunavukarasu, A.J.}, \bibinfo{author}{Ting, D.S.J.},
  \bibinfo{author}{Elangovan, K.}, \bibinfo{author}{Gutierrez, L.},
  \bibinfo{author}{Tan, T.F.}, \bibinfo{author}{Ting, D.S.W.},
  \bibinfo{year}{2023}.
\newblock \bibinfo{title}{Large language models in medicine}.
\newblock \bibinfo{journal}{Nature medicine} \bibinfo{volume}{29},
  \bibinfo{pages}{1930--1940}.
\bibitem[{Vassiliou et~al.(2023)Vassiliou, Alevizakis, Papadakis and
  Kondylakis}]{vassiliou2023isummary}
\bibinfo{author}{Vassiliou, G.}, \bibinfo{author}{Alevizakis, F.},
  \bibinfo{author}{Papadakis, N.}, \bibinfo{author}{Kondylakis, H.},
  \bibinfo{year}{2023}.
\newblock \bibinfo{title}{isummary: Workload-based, personalized summaries for
  knowledge graphs}, in: \bibinfo{booktitle}{European Semantic Web Conference},
  \bibinfo{organization}{Springer}. pp. \bibinfo{pages}{192--208}.
\bibitem[{Xu et~al.(2021)Xu, Wang, Chen and Zhao}]{xu2021prediction}
\bibinfo{author}{Xu, W.}, \bibinfo{author}{Wang, T.}, \bibinfo{author}{Chen,
  R.}, \bibinfo{author}{Zhao, J.L.}, \bibinfo{year}{2021}.
\newblock \bibinfo{title}{Prediction of initial coin offering success based on
  team knowledge and expert evaluation}.
\newblock \bibinfo{journal}{Decision Support Systems} \bibinfo{volume}{147},
  \bibinfo{pages}{113574}.
\bibitem[{Yang et~al.(2023)Yang, Liu, Ouyang, Ren and
  Wen}]{yang2023counterfactual}
\bibinfo{author}{Yang, Z.}, \bibinfo{author}{Liu, Y.}, \bibinfo{author}{Ouyang,
  C.}, \bibinfo{author}{Ren, L.}, \bibinfo{author}{Wen, W.},
  \bibinfo{year}{2023}.
\newblock \bibinfo{title}{Counterfactual can be strong in medical question and
  answering}.
\newblock \bibinfo{journal}{Information Processing \& Management}
  \bibinfo{volume}{60}, \bibinfo{pages}{103408}.
\bibitem[{Yang et~al.(2021)Yang, Xu and Chen}]{yang2021deep}
\bibinfo{author}{Yang, Z.}, \bibinfo{author}{Xu, W.}, \bibinfo{author}{Chen,
  R.}, \bibinfo{year}{2021}.
\newblock \bibinfo{title}{A deep learning-based multi-turn conversation
  modeling for diagnostic q\&a document recommendation}.
\newblock \bibinfo{journal}{Information processing \& management}
  \bibinfo{volume}{58}, \bibinfo{pages}{102485}.
\bibitem[{Yao et~al.(2023)Yao, Peng, Mao and Luo}]{yao2023exploring}
\bibinfo{author}{Yao, L.}, \bibinfo{author}{Peng, J.}, \bibinfo{author}{Mao,
  C.}, \bibinfo{author}{Luo, Y.}, \bibinfo{year}{2023}.
\newblock \bibinfo{title}{Exploring large language models for knowledge graph
  completion}.
\newblock \bibinfo{journal}{arXiv preprint arXiv:2308.13916} .

\end{thebibliography}

\end{document}